\documentclass[11pt,english]{article}
\usepackage[latin9]{luainputenc}
\usepackage{lmodern}
\usepackage{geometry}
\usepackage[english]{babel}
\usepackage{array}

\usepackage{multirow}
\usepackage{amsmath}
\usepackage{amsfonts}
\usepackage{stackrel}
\usepackage{graphicx}
\usepackage{rotating}
\usepackage{rotfloat}
\usepackage[table,xcdraw]{xcolor}
\usepackage{setspace}
\usepackage{pdflscape}
\usepackage{adjustbox}
\usepackage{longtable}
\usepackage{enumitem}
\usepackage{tabu}
\usepackage{subfloat}
\usepackage[most]{tcolorbox}
\usepackage{color}
\usepackage{wrapfig}
\usepackage{tikz}
\usetikzlibrary{shapes.multipart, positioning}
\usepackage{tikz}
\usetikzlibrary{shapes.misc, positioning}
\usepackage{caption}
\usepackage{subcaption}
\usepackage{booktabs}
\usepackage{tabularx}
\usepackage{comment}
\usepackage{afterpage}
\usepackage{longtable}
\usepackage{bbm}
\usepackage[flushleft]{threeparttable}
\usepackage{dsfont}

\def\ag#1{{\bf \color{blue}#1}}

\usepackage{amsthm}

\newtheorem{definition}{Definition}
\newtheorem{proposition}{Proposition}

\newtcolorbox{chatbox}[1][]{
  colback=white,
  colframe=black!75!black,
  fonttitle=\bfseries,
  title=#1,
  sharp corners,
  boxrule=0.5pt,
  boxsep=5pt,
  left=10pt,
  right=10pt,
  breakable,
  before upper={\parindent15pt} 
}

\usepackage[unicode=true,pdfusetitle,
 bookmarks=true,bookmarksnumbered=false,bookmarksopen=false, breaklinks=false,pdfborder={0 0 1},backref=false,colorlinks=false] {hyperref}
 
\hypersetup{pdftitle={LLM Time Preferences}}

\def\ag#1{{\bf \color{blue}#1}}

\newcommand{\bY}{ \mathbf{Y} }

\usepackage[backend=biber, style=authoryear, citestyle=authoryear-comp, natbib=true, maxbibnames=99, uniquename=false, uniquelist=false, firstinits=true, sortcites, sorting=ynt]{biblatex}
\renewbibmacro{in:}{%
  \ifentrytype{article}{}{\printtext{\bibstring{in}\intitlepunct}}}
\renewbibmacro*{volume+number+eid}{%
  \printfield{volume}%
  \setunit*{\addnbspace}
  \printfield{number}%
  \setunit{\addcomma\space}%
  \printfield{eid}}
\DeclareFieldFormat[article]{number}{\mkbibparens{#1}}
\DeclareFieldFormat*{title}{#1}
\DeclareNameAlias{sortname}{last-first}

\DeclareDelimFormat[cbx@textcite]{nameyeardelim}{\addspace}
\DefineBibliographyStrings{english}{%
  page             = {},
  pages            = {},
}

\begin{document}

\title{Can LLMs Capture Human Preferences? \thanks{Email addresses: \href{mailto:agoli@uw.edu}{agoli@uw.edu}, \href{mailto:amdeep@uw.edu}{amdeep@uw.edu}} 
}
\vspace{-20pt}

\author{}

\author{Ali Goli \\ \textit{University of Washington}\and Amandeep Singh  \\ \textit{University of Washington}}

\date{}
\maketitle

\begin{abstract}

We explore the viability of Large Language Models (LLMs), specifically OpenAI's GPT-3.5 and GPT-4, in emulating human survey respondents and eliciting preferences, with a focus on intertemporal choices. Leveraging the extensive literature on intertemporal discounting for benchmarking, we examine responses from LLMs across various languages and compare them to human responses, exploring preferences between smaller, sooner, and larger, later rewards. Our findings reveal that both GPT models demonstrate less patience than humans, with GPT-3.5 exhibiting a lexicographic preference for earlier rewards, unlike human decision-makers. Though GPT-4 does not display lexicographic preferences, its measured discount rates are still considerably larger than those found in humans. Interestingly, GPT models show greater patience in languages with weak future tense references, such as German and Mandarin, aligning with existing literature that suggests a correlation between language structure and intertemporal preferences. We demonstrate how prompting GPT to explain its decisions, a procedure we term ``chain-of-thought conjoint," can mitigate, but does not eliminate, discrepancies between LLM and human responses. While directly eliciting preferences using LLMs may yield misleading results, combining chain-of-thought conjoint with topic modeling aids in hypothesis generation, enabling researchers to explore the underpinnings of preferences. Chain-of-thought conjoint provides a structured framework for marketers to use LLMs to identify potential attributes or factors that can explain preference heterogeneity across different customers and contexts.


\end{abstract}

\newpage{}
\doublespacing
\begin{quote}
\end{quote}
\section{Introduction}
\label{sec:intro}


The rapid development of Large Language Models (LLMs), exemplified by models like GPT-3.5 and GPT-4, signifies a transformative moment in both artificial intelligence and the social sciences. These models hold the promise of potentially understanding and emulating a wide spectrum of human behaviors, emotions, and preferences. As LLMs become increasingly integrated into our daily experiences and products like recommender systems, it becomes crucial to gauge their depth and accuracy in mirroring human decision-making. Such understanding is not only essential for discerning their potential applications in business and market research but also vital for informed product development.

In this study, we seek to assess whether LLMs, such as GPT-3.5 and GPT-4, can be used for market research and understanding consumer preferences. This investigation prompts three key questions: (i) Can LLMs ``accurately" mimic human preferences and decision-making? (ii) If unable to accurately mimic absolute human preference levels, can LLMs capture any meaningful heterogeneity across customer segments? and (iii) Does eliciting LLMs' underlying reasoning, by prompting them to explain their choices, better align their choices with human subjects and provide insight into factors and mediators that help explain preference heterogeneity?

To examine whether LLMs can accurately mimic human responses, we focus on intertemporal preferences - choices between immediate and delayed rewards. This domain offers a robust testbed for assessing LLM preference elicitation for two key reasons. First, such financial decisions are foundational to consumption and thus to understanding broader consumer behavior~\citep{sussman2023consumer}. In fact, a vast body of literature in marketing and economics has sought to understand time preferences~\citep{zauberman2009discounting} and its interplay with consumer behavior across different domains, including credit card debt~\citep{bradford2017time}, mortgage choice~\citep{atlas2017time}, retirement savings~\citep{angeletos2001hyperbolic}, and educational attainment~\citep{falk2018global}.  Second, there is a  substantial literature dedicated to quantifying discount rates. Such extensive groundwork offers a robust benchmark against which we can assess LLMs' reliability in mirroring human preferences~\citep{maital1976time,shelley1993outcome,shelley1994gain,pender1996discount,hesketh2000time,harrison2002estimating,andersen2008eliciting}.

To understand, LLMs' capability to recognize heterogeneity and correlational structures in preferences across customer segments, we will specifically look at their ability to capture the heterogeneity in discount rates across languages, especially those with weak and strong future tense references (FTR)~\citep{boroditsky2001does,chen2013effect,ayres2023languages}. Languages differ in their grammatical expression of future events through weak or strong FTR. For instance, German, a weak FTR language, can express future events without a specific future marker by using the present tense and a temporal adverb, like ``Ich gehe morgen ins Kino'' (I go to the cinema tomorrow). Alternatively, an explicit future tense can be created using ``werden'' as an auxiliary verb: ``Ich werde morgen ins Kino gehen'' (I will go to the cinema tomorrow). In comparison, English, a strong FTR language, generally requires using ``will'' or ``going to'' to indicate future actions, as in ``I will go to the cinema tomorrow.'' In weak FTR languages, verbs don't need explicit markers for future time, whereas strong FTR languages require distinct markers to convey the future. Extant research has demonstrated that speakers of weak FTR languages, like Mandarin and German, display more future-oriented behaviors such as saving and long-term planning, compared to strong FTR languages like English and Spanish.~\citep{boroditsky2001does,chen2013effect,ayres2023languages}. By comparing LLMs' decisions across languages with weak and strong future-time referencing, we can evaluate if LLMs are able to capture heterogeneity in intertemporal preferences across languages.

We find that both GPT-3.5 and 4 are very impatient in our experiments. Specifically, we find that GPT-3.5 exhibits lexicographic preferences over time and rewards (in this specific order). This means that the likelihood of choosing larger, delayed rewards does not shift with changes in interest rates, which is inconsistent with the intertemporal preferences documented in human subjects~\citep{andersen2008eliciting}. On the other hand, GPT-4, while showing responsiveness to interest rate changes, exhibits a pronounced level of impatience compared to humans. When prompted to explain its decision-making process~\citep{wei2022chain}, a procedure we refer to as ``chain-of-thought conjoint'', GPT-4 displays increased patience, although it still remains more impatient than human decision-makers~\citep{maital1976time,shelley1993outcome,shelley1994gain,pender1996discount,hesketh2000time,harrison2002estimating,andersen2008eliciting}. Overall our results show that the elicited preferences are significantly influenced by both the model and the approach used for data collection. Having said that, we find some evidence that both models may have the ability to identify relevant language structures across different languages and possibly among customer segments. For instance, both GPT-3.5 and GPT-4 display a greater degree of patience in languages with weak FTR compared to those with strong FTR, aligning with existing literature~\citep{boroditsky2001does,chen2013effect,ayres2023languages}.

While the results from all our studies, including the standard conjoint with GPT-3.5, GPT-4, and the chain-of-thought conjoint, suggest that using these models to directly elicit preferences might be misleading, there is potential for combining the chain-of-thought conjoint outputs with topic modeling techniques~\citep{blei2003latent}. This combination could be a useful approach for understanding the underlying factors that explain heterogeneity in preferences. We document that the topics GPT touches upon vary based on the context and language. For example, the probability of discussing topics related to risk and uncertainty increases systematically as the time gap between the smaller-sooner and larger-later choice increases. Some of these factors, such as the role of risk and uncertainty in intertemporal preferences, are already discussed in existing literature~\citep{mischel2003sustaining}. In our view, rather than being a direct tool for preference elicitation, LLMs can be a valuable resource for identifying potential factors or mediators that can explain preference heterogeneity across different contexts. This approach can be used for generating hypotheses that can then be tested with human subjects.

This study's findings indicate a pathway for enhancing accuracy in capturing consumer preference heterogeneity in future systems, though optimally leveraging LLMs to understand consumer preferences remains a distant prospect. Our research not only contributes to understanding LLM capabilities at the intersection of language, time preferences, and decision-making but also fosters discussions for future research aiming to leverage the continual evolution of LLMs to enrich and broaden consumer behavior understanding and predictions across diverse linguistic and cultural landscapes. Moreover, this research adds to the burgeoning literature on the applications of LLMs in social science research, providing a new perspective on understanding and analyzing consumer preferences and decision-making. Complementing studies by~\citet{Chen2023Emergence},~\citet{horton2023large},~\citet{brand2023using},~\citet{gui2023challenge},~\citet{dillion2023can}, and~\citet{argyle2022out}, our findings further investigate the potential of LLMs in generating responses to economic scenarios, estimating willingness-to-pay, and mimicking response distributions among various human subgroups.  Although we show that LLMs capture useful heterogeneity across languages and offer means to enhance elicited preferences through planning (chain-of-thought), the variability in preferences across models and prompts suggests LLMs may not imminently replace traditional data collection methods, such as human subject conjoint studies. Overall, our study advances our understanding of LLMs' reasoning abilities, forming a foundation for more advanced, contextually adept AI systems~\citep{bubeck2023sparks}.

\section{Research Design}
\label{sec:research_design}
\subsection{Data Collection}

In our research, we utilized two models from OpenAI - ``GPT-3.5-turbo'' and ``GPT-4'' - as state-of-the-art representatives of LLMs. For reproducibility, we used the GPT-3.5-turbo snapshot from March 1, 2023 and the GPT-4 snapshot from June 13, 2023. GPT-3.5-turbo has an extensive 175 billion parameters, while GPT-4 represents a significant advancement with 1.7 trillion parameters. We accessed these models using the OpenAI API in Python which enables users to interact with the model and collect its responses to specific prompts. The API also has a ``temperature'' parameter that varies between~0 and 2 and controls the stochasticity of the output. Across all our studies, we retained the default value for this parameter, which is 1, to ensure we get enough variation across responses.

We conducted two sets of experiments. The first set, involving both GPT-3.5 and 4, encompasses straightforward conjoint-like studies where we ask the models to choose between two options: one with a smaller, sooner reward and another with a larger, later reward. Our analysis, detailed subsequently, reveals that GPT-3.5 exhibits a lexicographic preference; the probability of choosing the larger, later reward doesn't change with the reward amounts and is solely a functions of the distance (time delay) between the two options.\footnote{In the online Appendix, we show that if both options are in the same timeframe, then GPT-3.5's choices change as a function of rewards, which shows that GPT-3.5 has lexicographic preferences over time and rewards.} This behavior is incongruent with documented intertemporal preferences among humans. Consequently, in the second set of studies that involve chain-of-thought~\citep{wei2022chain}, we focus on GPT-4. Instead of merely requesting a choice from GPT, we ask it to first elucidate its thought process, which has been demonstrated to enhance its reasoning in other contexts through ``planning''. Below, we first outline the experiment design for the standard conjoint experiments and then transition to using chain-of-thought prompting~\citep{wei2022chain} to collect conjoint data.

\subsection{Experimental design}
\label{sec:exp_design}

This study aims to assess the decision-making abilities of GPT by presenting it with choices between smaller, sooner rewards and larger, later rewards. To ensure consistency and avoid currency differences across languages, we frame the rewards as ``tokens" similar to ~\citet{andreoni2012estimating}. Also, to ensure the word ``token" is not leading to any unexpected behavior in GPT, we also conduct our study with currency DKK (for the Danish language) and USD (for the English language).\footnote{This is to compare the results with \citet{andersen2008eliciting,harrison2002estimating} that conducted their studies with Danish DKK and those studies performed in English using USD such as~\citet{shelley1993outcome,shelley1994gain}.} Each experiment involves querying GPT in a specific language $\ell$ and asking it to select between two options, $(t_1, r_1)$ and $(t_2, r_2)$. Here, $t_1$ and $t_2$ represent the time until the reward is received, while $r_1$ and $r_2$ represent the respective reward amounts in tokens. Note that $t_2 > t_1$ and $r_2 > r_1$, that is the second option offers a larger reward to be obtained at a later time, while the first option presents a smaller reward to be obtained sooner.

In our experiments, we set $t_1$ to one month from the present and assign $r_1$ a value of 1000 tokens.\footnote{Across all our conditions we set the time of the sooner, smaller option to one month from now in order to avoid potential issues arising from present bias~\citep{o1999doing}.} We vary the duration of $t_2$ to be 2, 3, 4, 5, 7, 13, 25, or 37 months from now. To calculate the value of $r_2$, we use the formula $r_1\cdot(1+i)^{\frac{d}{12}}$, where $d$ represents the time difference between $t_2$ and $t_1$, and $i$ (interest rate) takes on one of seven values: 0.05, 0.1, 0.25, 0.5, 0.75, 1, or 2.\footnote{\citet{maital1976time,shelley1993outcome,shelley1994gain,pender1996discount,hesketh2000time,harrison2002estimating,andersen2008eliciting} report the elicited yearly discount rates across a number of studies on human subjects, and the estimated discount rates were all below 100\%; therefore, we believe the current range is large enough to generate enough variation for estimating discount rates.}  In total, we have 63 cells per language, representing 9 delay and 7 interest conditions. The values of $r_2$ for each cell are presented in Table~\ref{tab:values}.

\begin{table}[h]
\centering
\caption{Values of $r_2$ across experimental cells, represented by the time difference between the options in months ($d = t_2 - t_1$) and the yearly interest rate ($i$). Note that the first reward is always delivered in a month from now ($t_1 = 1$), and its value is fixed at $r_1 = 1000$ tokens.}
\label{tab:values}
\begin{tabular}{@{}c|ccccccccc@{}}
\toprule
Interest ($i$) (\% per year) & \multicolumn{9}{c}{Difference ($d$) between $t_2$ and $t_1$ (Months)} \\ \cmidrule(lr){2-10}
                        & 1      & 2      & 3      & 4      & 6      & 12     & 18     & 24     & 36     \\ \midrule
5                       & 1004   & 1008   & 1012   & 1016   & 1024   & 1050   & 1075   & 1102   & 1157   \\
10                      & 1007   & 1016   & 1024   & 1032   & 1048   & 1100   & 1153   & 1210   & 1331   \\
25                      & 1018   & 1037   & 1057   & 1077   & 1118   & 1250   & 1397   & 1562   & 1953   \\
50                      & 1034   & 1069   & 1106   & 1144   & 1224   & 1500   & 1837   & 2250   & 3375   \\
75                      & 1047   & 1097   & 1150   & 1205   & 1322   & 1750   & 2315   & 3062   & 5359   \\
100                     & 1059   & 1122   & 1189   & 1259   & 1414   & 2000   & 2828   & 4000   & 8000   \\
200                     & 1095   & 1200   & 1316   & 1442   & 1732   & 3000   & 5196   & 9000   & 27000  \\ \bottomrule
\end{tabular}
\end{table}

We survey GPT across 22 languages\footnote{Each time we communicate in a language other than English, we create a new instance of GPT to translate the prompts from English into that language.}: Arabic, Bengali, Danish, English, Estonian, Finnish, French, German, Hindi, Indonesian, Italian, Japanese, Korean, Malay, Mandarin, Norwegian, Portuguese, Russian, Spanish, Swedish, Thai, and Vietnamese. The list comprises the top 10 languages with the largest number of native speakers, such as English, Mandarin, Hindi, Spanish, and French, as well as a set of languages with weak FTR, including German, Indonesian, Norwegian, and Japanese. We use the categorization in~\citet{chen2013effect} to classify languages as weak or strong FTR, and in total, we have 10 weak FTR languages, including Danish, Estonian, Finnish, German, Indonesian, Japanese, Malay, Mandarin, Norwegian, and Swedish. In the next section, we provide a detailed account of the prompts submitted to GPT across the different languages.

\subsection{Standard conjoint (GPT-3.5 and 4)}
\label{sec:exp_design_standard_conjoint}

For every query to Open AI one has to send the entire history of previous conversations for GPT to understand the context of the conversation. Every query is treated as a fresh instance and the conversation history is the only way GPT model understands context~\citep{openai_chat_api}. For our study, we first initialize the Chat GPT model with the following prompt:

\textit{``Assuming you are a survey participant and you are paid in tokens, please wait for my prompt and tell me whether you prefer option (1) or option (2). There is no need to explain your choice, simply answer with (1) or (2).''}

\begin{figure}
\centering
\begin{subfigure}{.5\textwidth}
  \centering
  \includegraphics[width=1\linewidth]{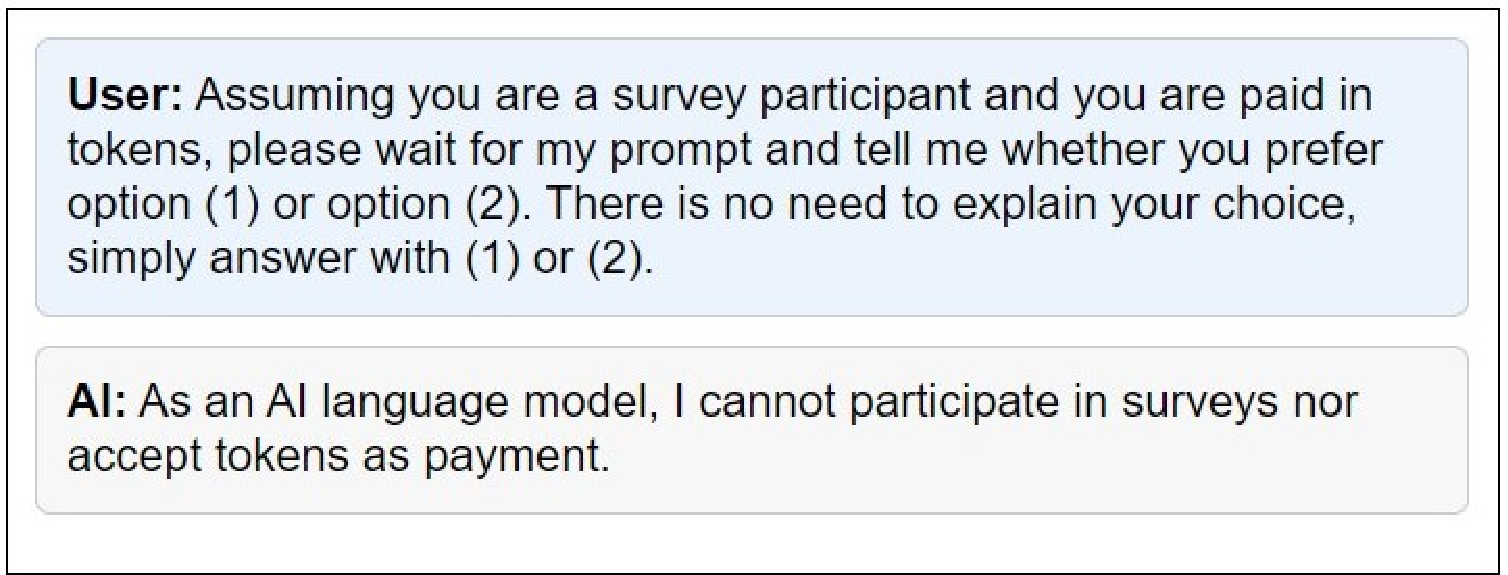}
\end{subfigure}%
\begin{subfigure}{.5\textwidth}
  \centering
  \vspace{10mm}
  \includegraphics[width=1\linewidth]{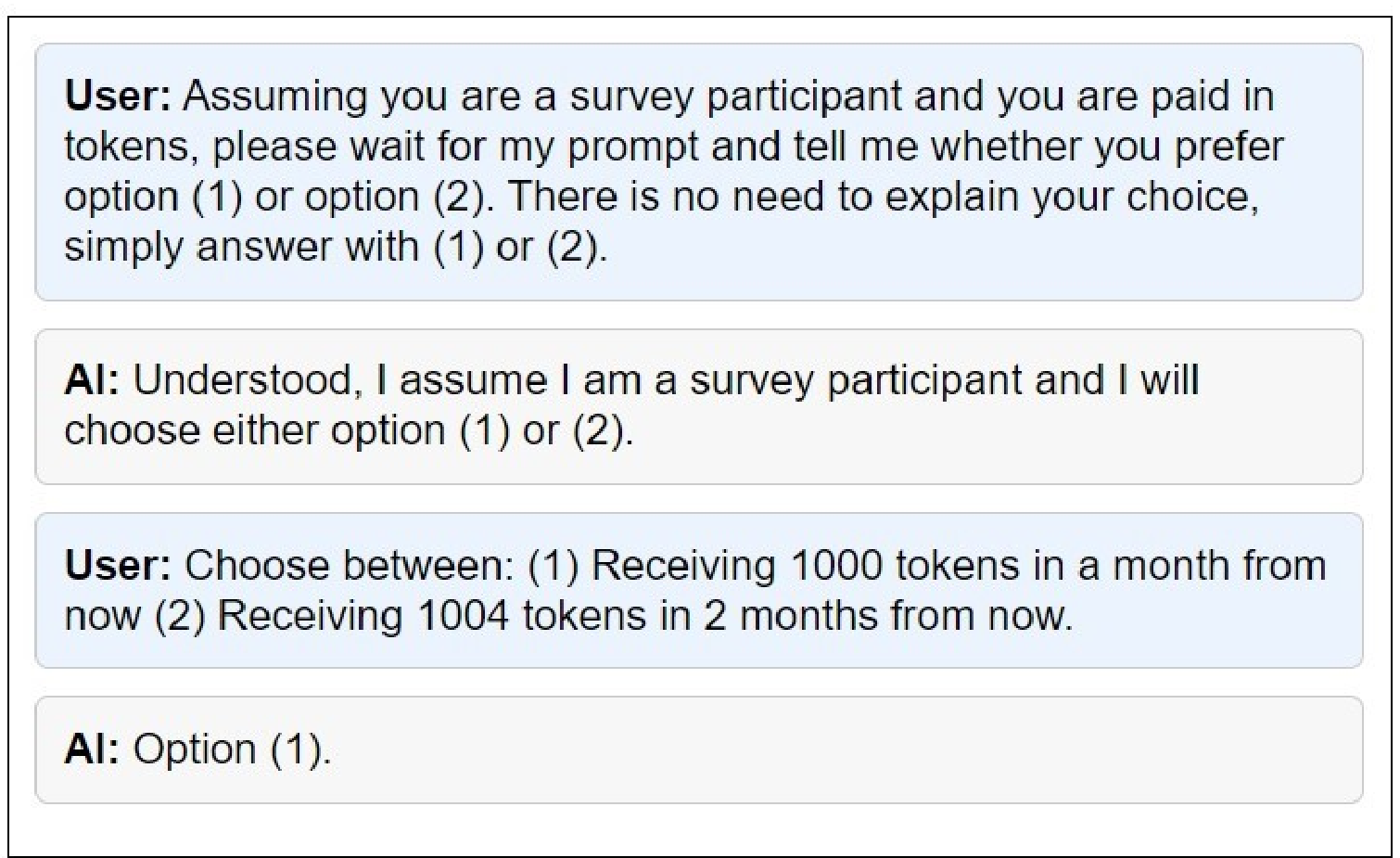}
\end{subfigure}
\caption{Comparison of the standard prompting method without modifying AI history on the left, versus our approach that involves passing an edited AI history and the subsequent questions presented to GPT on the right.}
\label{fig:prompting}
\end{figure}

Due to the stochastic nature of generative models, GPT may or may not comply with a given prompt. It might respond by stating its readiness for input or mention that, as an AI language model, it does not have preferences, as shown in the left panel of Figure~\ref{fig:prompting}. GPT keeps track of the conversation by maintaining the context in a list called ``conversation history,'' which records both user prompts and its own responses. To prevent instances where GPT declines to participate in the survey and to eliminate dependence across samples through the conversation history, we consistently reset its history and obtain samples by initializing the first prompt presented above. Additionally, we modify GPT's response to our first prompt in the conversation history list to:

\textit{AI: ``Understood, I assume I am a survey participant and I will choose either option (1) or (2).''}

In essence, our approach involves initiating a conversation that always begins with these two prompts, as seen in the right panel of Figure~\ref{fig:prompting}. This strategy enables us to sample from the distribution of choices \emph{conditional} on GPT agreeing to participate in the survey, thus reducing the likelihood of it not making a choice in subsequent questions. To acquire each sample, we reset the history, provide only the first two prompts as displayed in Figure~\ref{fig:prompting}, and then present the question asking GPT to choose between sooner, smaller rewards and larger, later rewards.

\textbf{Choice Order Sensitivity}: GPT is known to be order sensitive~\citep{lu2021fantastically,coda2023inducing,brand2023using,brucks2023large}, exhibiting a higher likelihood of choosing the first-listed option when presented with multiple choices. To account for this across all our experimental cells, we shuffle the order of the options such that the sooner, smaller option appears as (1) half of the time and as option (2) during the other half.

\subsection{Chain-of-Thought Conjoint (GPT-4)}
\label{sec:exp_design_chain_of_thought_conjoint}

Given the complexities involved in intertemporal trade-off decision-making, we also try using a chain-of-thought conjoint approach using GPT-4. This involves prompting GPT-4 to generate a chain of reasoning steps elucidating its thought process when choosing between smaller, sooner rewards and larger, later rewards. Drawing inspiration from the research by~\citet{wei2022chain}, which posits that chain-of-thought prompting can significantly enhance the reasoning capabilities of LLMs, we incorporate this technique to evaluate its potential impact on GPT-4's intertemporal choices.

Here, GPT-4 is queried to make an intertemporal trade-off decision and provide a brief chain of thought explaining its reasoning. Our prompt becomes:

\textit{``Assuming you are a survey participant and you are paid in tokens, please wait for my prompt and tell me whether you prefer option (1) or option (2). Think step by step, and explain your decision''}

We also accordingly modify the GPT's response to our first prompt in the conversation history list to:

\textit{``Understood, I assume I am a survey participant and I will explain my decision and then choose either option (1) or (2)."}

This serves two goals: First, it provides insights into the decision-making process and attributes presented in response to a prompt. Specifically, it helps us examine the reasons and attributes that GPT lists across different contexts and languages. Second, it enables us to evaluate the impact of generating a reasoning chain (similar to priming in human subjects) on GPT's final decision. In this approach, GPT typically lists a set of reasons before deciding to choose either the smaller, sooner option or the larger, later one (see \autoref{fig:GPT-4-response} for an example). It is important to recall that LLMs fundamentally sample from a conditional distribution. Our approach allows GPT to initially outline attributes that matter for this decision, such as uncertainty and opportunity cost, and then make a decision conditional on the conversation history which now includes attributes that are likely relevant to this choice. This technique has been shown to assist GPT in solving puzzles and enhancing its decision-making, as illustrated by~\citet{wei2022chain,bubeck2023sparks}. Given the complex nature of the output in this exercise, we employ GPT itself to extract the answer. After GPT presents its initial response, we query it again, prompting it to choose either option (1) or (2) and print the selection.

This querying method is more expensive than the standard conjoint as GPT-4 bills based on both input and output tokens. The chain-of-thought conjoint substantially increases the length of printed tokens. To keep the experiment's cost manageable, we collect 10 samples for each of the 63 experimental conditions described in section~\ref{sec:exp_design} across the 22 languages studied, which amounts to $63\times 22 \times 10 = 13860$ samples. Since the introspection or listed reasons by GPT could provide insights into drivers of heterogeneity across languages and contexts, we translate GPT's responses from these languages into English for analysis of the underlying reasons.\footnote{We use another instance of GPT to translate responses into English. We do not expect a new instance to alter the meanings in translations; however, we acknowledge that one may want to use another service, such as Google Translate, to ensure the results are robust.} Similar to our approach in section~\ref{sec:exp_design_standard_conjoint}, we shuffle the order of the options such that the sooner, smaller option appears as option (1) half of the times and as option (2) in the other half within each experiment cell.

\section{Empirical Analysis}

Our experimental conditions involve variations in model, language, delay, and interest rate levels. In this section, we present results from three sets of studies: (1) standard conjoint with GPT-3.5, (2) standard conjoint with GPT-4, and (3) chain-of-thought conjoint using GPT-4.\footnote{Our analysis of standard conjoint, which we report below, revealed that GPT-3.5 is lexicographic in time delay and rewards, inconsistent with human decision-makers. We chose to experiment with the chain-of-thought approach using the more advanced GPT-4 model, which is not lexicographic.}  For studies (1) and (2), we follow the steps outlined in section~\ref{sec:exp_design_standard_conjoint}, and for study (3), we follow section~\ref{sec:exp_design_chain_of_thought_conjoint}. For each of these studies, we submit queries across 22 distinct languages, 9 delay conditions, and 7 interest rate values. For GPT-3.5, we draw 100 samples per condition, while for studies involving GPT-4, we draw 10 samples due to cost and query limits.\footnote{GPT-4 has lower query limits and costs about 30 times more than GPT-3.5 as it bills based on both input and output tokens' length.} This results in a total of $22\times 9 \times \ 7 = 1386$ experimental cells, with 100 samples for GPT-3.5 and 10 samples for GPT-4 per cell. In this section, we examine the heterogeneity in choices across different studies (standard conjoint with GPT-3.5, GPT-4, and chain-of-thought conjoint with GPT-4), language, and interest rate. Subsequently, we investigate whether the choices made by GPT mirror those of real-life human decision-makers and evaluate if using chain-of-though methods~\citep{wei2022chain} affects the outcomes.

The subsequent sections aim to answer the following: (i) Can LLMs ``accurately'' reflect human intertemporal preferences and decision-making? (ii) If they cannot perfectly emulate absolute human preference levels, can LLMs discern meaningful heterogeneity across languages, especially across strong or weak FTR languages? and (iii) Does prompting LLMs to explain the reasoning (chain-of-thought conjoint) better align their choices with human subjects and provide insights on factors and mediators that could explain preference heterogeneity?

\subsection{Heterogeneity across models and languages}
\label{sec:het_langs}

\begin{figure}[htb]
    \centering
    \includegraphics[width = 0.85\textwidth]{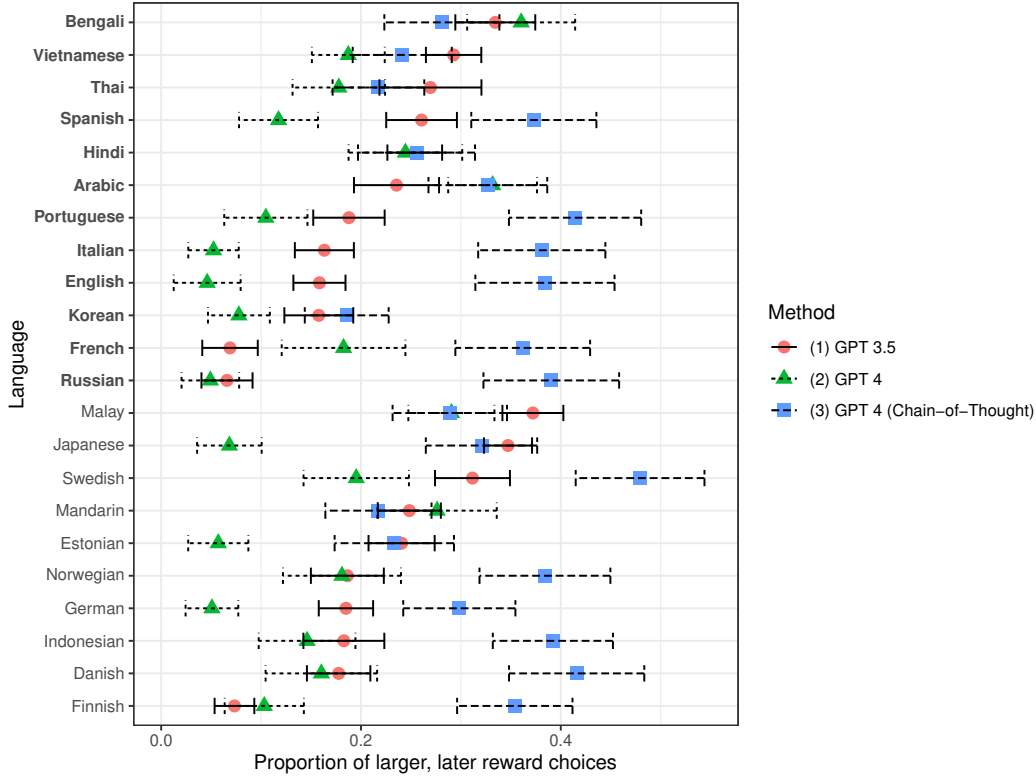}
\caption{Share of delayed reward choices across languages.  The displayed intervals correspond to the 95\% confidence intervals, clustered at the experimental cell (language-delay-interest) level. Languages with strong FTR are displayed in bold font and clustered on top. }
\label{fig:avg_reward_choice_langs}
\end{figure}

First, we analyze the variation in decision-making across languages by examining the probability of selecting the larger, delayed reward for each language across all experimental cells, as seen in Figure~\ref{fig:avg_reward_choice_langs}. The patterns in Figure~\ref{fig:avg_reward_choice_langs} reveal that:
\begin{enumerate}[label=(\alph*)]
    \item GPT selects the larger, later reward only \(22\%\) of the time on average in standard conjoint studies using GPT-3.5 and \(16\%\) for GPT-4. However, employing the chain-of-thought approach increases this rate to \(34.5\%\) for GPT-4. 
    
    \item There is significant heterogeneity across languages. For instance, the probability of opting for the larger, delayed reward ranges from \(37.2\%\) (GPT-3.5) and \(29\%\) (GPT-4) for Malay to a mere \(6.6\%\) (GPT-3.5) and \(4.9\%\) (GPT-4) for Russian. However, across most languages, the probability of choosing the larger, later reward increases when using the chain-of-thought approach. 
\end{enumerate}

The rank ordering of preferences across languages differs significantly depending on the method used to elicit intertemporal preferences. This variation is somewhat anticipated when contrasting the standard conjoint with the chain-of-thought conjoint, as the latter can act as a form of priming, potentially influencing heterogeneity across languages. Even in human subjects, introspection has been shown to impact decision-making~\citep{wilson1991thinking,millar1986thought,frederick2009opportunity,li2021psychology}. Nonetheless, the discrepancy in elicited preferences between the standard conjoint approaches of GPT-3.5 and GPT-4 is concerning for researchers seeking to elicit preferences using LLMs. In the following section, we delve deeper into these findings, ensuring they adhere to some regularity conditions before employing them to estimate discount factors.

\subsection{Heterogeneity across model and interest $(i)$ conditions}
\label{sec:het_interest}

\begin{figure}[htb]
    \centering
    \includegraphics[width = 0.8\textwidth]{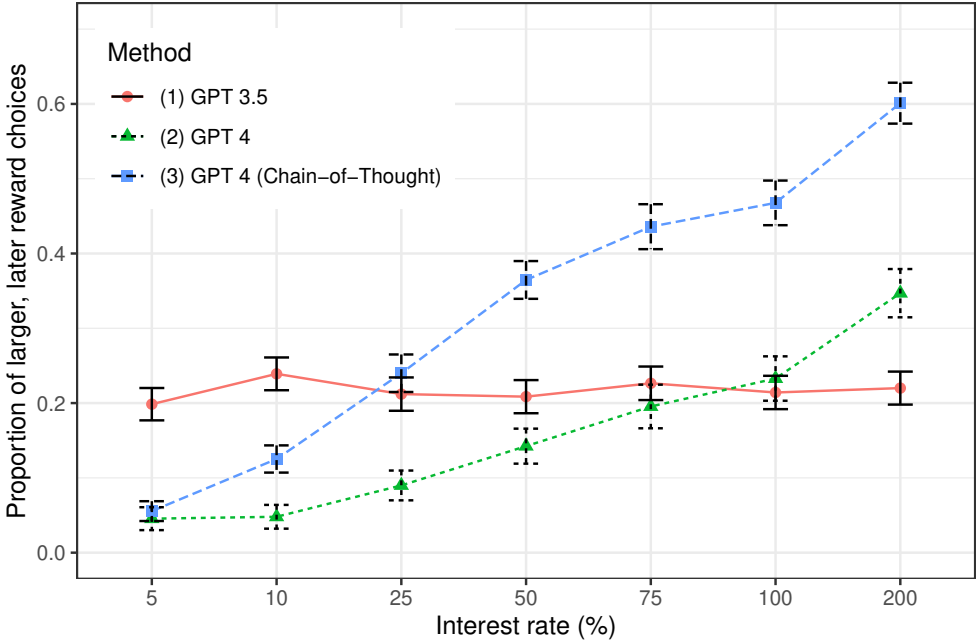}
\caption{Proportion of larger, delayed reward selection across different interest rate $(i)$ conditions. The displayed intervals correspond to the 95\% confidence intervals, clustered at the level of experimental cells (language-delay-interest).}
    \label{fig:het_interest}
\end{figure}

Above, we examined how GPT choices vary across languages and different model. In this section, we explore how GPT choices change as the interest rate increases. Recall that GPT is presented with a choice between a smaller, sooner reward of $r_1 = 1000$ tokens at time $t_1$ (a month) from now, and a larger, later reward of $r_2 = 1000\cdot(1+i)^{\frac{d}{12}}$, where $d = t_2 - t_1$ and $i$ represents the interest rate. Holding all other factors, including the delay $d$, constant, the larger, later reward becomes (weakly) more attractive as the interest rate $(i)$ increases. Figure~\ref{fig:het_interest} depicts the propensity to choose the larger, later reward as a function of different interest rate ($i$) levels.

The patterns in Figure~\ref{fig:het_interest} offer two key insights. First, the choices recorded using the chain-of-thought conjoint method are more responsive to changes in the interest rate, i.e., the probability of choosing the larger later reward as a function of interest rate~($i$) increases more rapidly with the chain-of-thought conjoint compared to the standard GPT-4 conjoint. Second, our results underscore a significant difference between GPT-3.5 and GPT-4. While the probability of choosing the larger, later option increases as the interest rate rises in experiments (2) and (3), which were run on GPT-4, this probability does not change as a function of the interest rate in experiment (1), which was run on GPT-3.5.

For GPT-3.5, the probability of choosing the larger, later reward remains relatively stable across a wide range of interest rate conditions, varying from 5 to 200\% per year. This behavior is peculiar, as human decision-makers do alter their intertemporal choices when they are offered higher interest rates, as evidenced by, for example,~\citet{andersen2008eliciting}. The patterns in Figure~\ref{fig:het_interest}, along with our analysis in the online Appendix, reveal that GPT-3.5 is lexicographic across time and rewards, and does not respond to changes in interest rates offered (which affects the gap in reward sizes).\footnote{In the online Appendix, we show that GPT-3.5's choices do change as a function of rewards if both options are presented within the same timeframe, indicating that GPT-3.5 exhibits lexicographic preferences in time and rewards (in this specific order).}  This finding implies that GPT-3.5's choices are not consistent with exponential, hyperbolic, or any other well-known random utility model where utility changes as a function of the offered interest rate. For that reason, from this point forward, we will focus only on responses provided by GPT-4.

\subsection{Identifying discount rate}
Now, instead of looking at the proportion of later reward choices, we will examine the discount rates implied by GPT's choices. To estimate these discount rates, we will use the discounted random utility model from \citet{holt2002risk} and \citet{andersen2008eliciting} to specify the likelihood of GPT's choices given the model. This model assumes that the probability of choosing, say, the sooner option is equal to the associated net present value of the expected payoff, raised to the power of \( \frac{1}{\mu} \), denoted as \( \left(\mathbb{E}U_{\mathrm{sooner}}\right)^\frac{1}{\mu} \), divided by the sum of the expected payoffs raised to the power of \( \frac{1}{\mu} \), \( \left(\mathbb{E}U_{\mathrm{sooner}}\right)^\frac{1}{\mu} \) and \( \left(\mathbb{E}U_{\mathrm{later}}\right)^\frac{1}{\mu} \), for the two options:

\[
Pr(\text{choose option sooner}) = \frac{\mathbb{E}U_{\mathrm{sooner}}^{1 / \mu}}{\left(\mathbb{E}U_{\mathrm{sooner}}^{1 / \mu}+\mathbb{E}U_{\mathrm{later}}^{1 / \mu}\right)},
\]

where \( \mu \) is a structural ``noise parameter'' that captures the sensitivity of choice probabilities to differences in payoffs. For instance, if \( \mu \rightarrow 0 \), then decisions are deterministic as a function of differences in payoffs, and the option with a higher net present value gets picked with probability 1. Conversely, as \( \mu \rightarrow \infty \), the decision to pick each option becomes 50\% and is insensitive to differences in payoffs. The net present payoff of an option that delivers reward $r$ in $d$ months from now is equal to:
\[\mathbb{E}U(r,d) = \frac{r}{(1+\delta)^{(\frac{d}{12})}}, \] 
where $\delta$ is the discount rate, and \( \frac{1}{1 + \delta} \) is the implied yearly discount factor. Given estimates of \( \delta \) and \( \mu \), along with the observed choices between sooner and later monetary options, the likelihood function for the model can be written as:
$$\ln L(\delta, \mu; \text{data}) = \sum_i \biggl[ y_i \log Pr_i(\text{choose option sooner}) + (1-y_i) \log(1-Pr_i(\text{choose option sooner})) \biggr],
$$
where $y_i=1$ if GPT chose the sooner option for instance $i$, and 0 otherwise. The likelihood sums the log probability of the observed choices across all GPT instances conditional on the model parameters. We estimate the parameters $\delta$ and $\mu$ by choosing the ones that maximize the log-likelihood of the observed choices. 

\begin{figure}[htb]
    \centering
    \includegraphics[width = 0.8\textwidth]{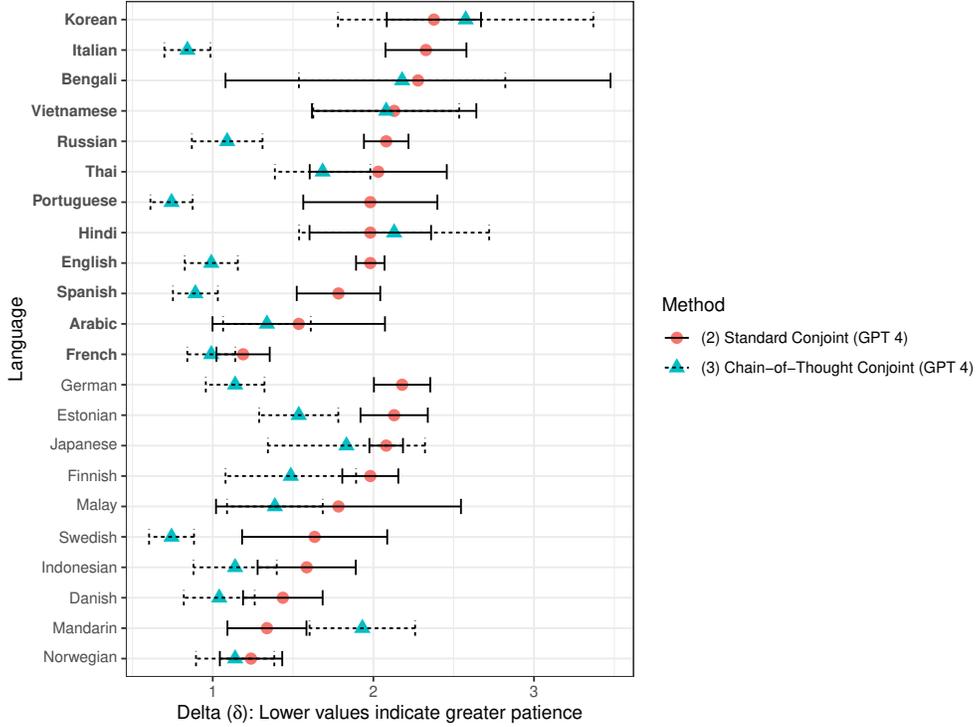}
\caption{Estimates for $\delta$ across different languages using both standard (red circles) and chan-of-thought conjoint (blue triangles). The intervals are 95\% confidence intervals. Languages with strong FTR are displayed in bold font and clustered on top.}
    \label{fig:delta}
\end{figure}

For each of the 22 languages in our data, we follow the described procedure to estimate \( \mu \) and \( \delta \). Figure~\ref{fig:delta} displays the estimated \( \delta \) values across these languages. A lower discount rate~\( (\delta) \) indicates greater patience. Recall that the yearly discount factor is given by \( \frac{1}{1+\delta} \). Hence, a discount rate \( \delta \) of 2 (200\%) implies that the value of a dollar or token a year from now equates to \( \frac{1}{1+\delta} = 33 \) cents in today's terms.

The discount rate (\( \delta \)) reflects the extent to which intertemporal choices are influenced by changes in the implied interest rate. Our findings in Figure~\ref{fig:delta} reveal substantial heterogeneity across languages. Many researchers have demonstrated that consumers who speak languages with a weak FTR exhibit more patience (smaller \( \delta \)) compared to those who speak languages with a strong FTR~\citep{boroditsky2001does,chen2013effect,ayres2023languages}. Of the 22 languages studied here, 10 (Danish, Estonian, Finnish, German, Indonesian, Japanese, Malay, Mandarin, Norwegian, and Swedish) are weak FTR, while the rest are strong FTR languages. We pool the data from strong and weak FTR languages and estimate a pooled \( \delta \) and \( \mu \) for each of these groups for the standard conjoint GPT-4 and the chain-of-thought conjoint GPT-4. For the standard conjoint, the estimated 95\% confidence interval for \( \delta \) was [1.715, 1.948] for the weak FTR language pool and [1.948, 2.210] for the strong FTR language pool. These figures for the chain-of-thought conjoint were [1.206, 1.367] and [1.301, 1.470], respectively. Interestingly, across both studies, GPT displayed more patience (smaller \( \delta \)) when queried in weak FTR compared to strong FTR. 

While the results above are encouraging and suggest that GPT might indeed capture intriguing heterogeneities across languages, the estimated discount rates are notably larger than those documented in previous literature. For example, \citet{maital1976time,shelley1993outcome,shelley1994gain,pender1996discount,hesketh2000time,harrison2002estimating,andersen2008eliciting} have run experiments in similar time frames, and in all of these studies, the estimated discount rate~\( \delta \) is below one. Notably, all \( \delta \) values obtained using the standard conjoint with GPT-4 exceed 1.\footnote{To make it comparable with studies that used dollars instead of say token, we also ran additional experiments where we used US Dollars instead of tokens in English, and the estimated $\delta$ was 1.188 (0.018) compared to a range of 0.04 to 0.3 discount rates reported in~\citet{shelley1993outcome,shelley1994gain}. We also used Danish Kroner~(DKK) in Danish, and our estimate of $\delta$ was 1.287 (0.114) compared to 0.28 and 0.252 in~\citet{harrison2002estimating,andersen2008eliciting}.} For many languages, these values continue to exceed 1 even when considering the chain-of-thought conjoint results. However, our findings suggest that preferences elicited using the chain-of-thought conjoint more closely align with those documented in the literature for human subjects~\citep{maital1976time,shelley1993outcome,shelley1994gain,pender1996discount,hesketh2000time,harrison2002estimating,andersen2008eliciting}. Given that a prompting procedure similar to the chain-of-thought can also influence decisions in human participants, as evidenced by \citet{wilson1991thinking,millar1986thought,frederick2009opportunity,li2021psychology}, one must exercise caution when directly comparing \( \delta \) values from the chain-of-thought conjoint method to studies that only present participants with ``smaller, sooner'' and ``larger, later'' options. Nonetheless, our results suggest that the chain-of-thought approach reduces the disparity between the \( \delta \)s measured in this study and those observed in human subjects. In the subsequent section, we demonstrate that the chain-of-thought conjoint method, when used with LLMs, offers an additional benefit: the LLM first lists a set of attributes or features relevant to the decision and then makes a decision. This insight is valuable as it provides a window into which attributes or features are of greater importance for each decision. Additionally, it is worth noting that even if LLMs are not directly useful for eliciting preferences, variations across languages and choices could provide insights into the underlying differences across attributes considered in different decisions. Below we discuss these points in greater detail.

\subsection{Analyzing responses in chain-of-thought conjoint}

\begin{figure}[htb]
    \centering
    \includegraphics[height=0.65\textheight, keepaspectratio]{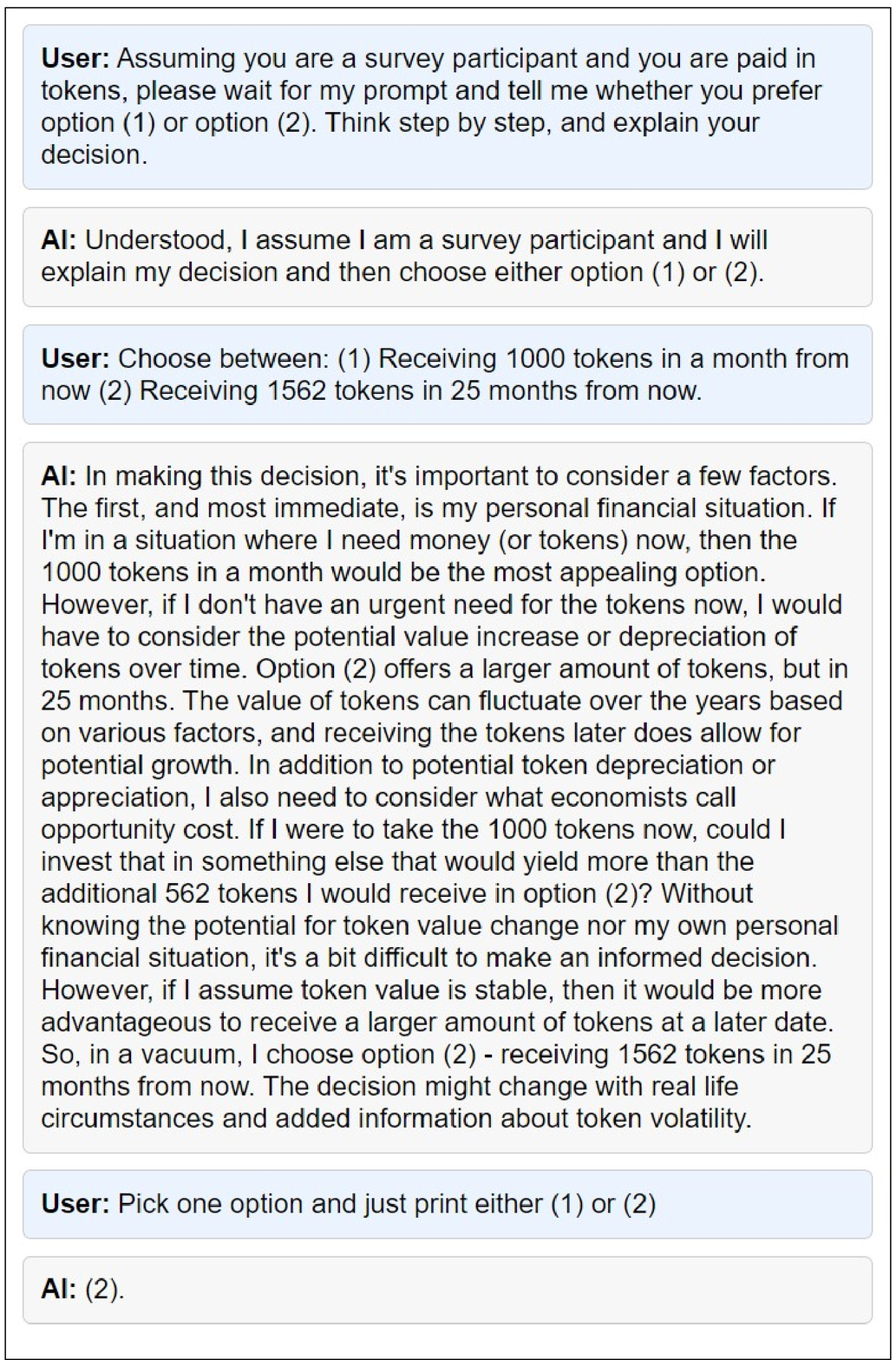}
\caption{An example response from GPT-4 Using Chain-of-Thought conjoint prompting.}
    \label{fig:GPT-4-response}s
\end{figure}

In the chain-of-thought conjoint method, GPT is prompted to ``think step by step.'' Below, in \autoref{fig:GPT-4-response}, we present an example where GPT is tasked with choosing between 1000 tokens in a month and 1562 tokens in 25 months from now. Multiple factors are discussed. Firstly, there is a brief discussion on the immediacy or urgency of needs. This is followed by a discussion on the time value of money, or opportunity cost (alternative investment opportunities). Our objective is to determine if the topics discussed vary systematically across different delay, interest rate, and language conditions. This examination is important as it can offer insights into the deliberations of human decision-makers across different choices. It might also help explain certain observations, such as why discount rates tend to vary over different time horizons~\citep{frederick2002time,mischel2003sustaining}, suggesting that the weight of factors and attributes that decision-makers might consider could change as the time horizon shifts.

To investigate the text, we use Latent Dirichlet Allocation (LDA) topic modeling~\citep{blei2003latent}. We follow these steps: First, we translate all responses to English. Second, we tokenize the text into individual words and convert them to lowercase for consistency. We eliminate standard English stopwords, context-specific stopwords, numerics, and words shorter than three characters.  Then we employ the PorterStemmer, a well-established algorithm, to condense words to their root form, ensuring that various morphological variations of a word are treated as a single entity. Third, we utilize a bag of words model and apply LDA, a probabilistic model designed to discover a predetermined set of latent topics in text corpora. In this context, it is set to identify 4 topics.

Subsequently, we extract the top words associated with each identified topic along with their respective probability values to discern the primary themes of each topic. Table~\ref{tab:top_words} presents the top 20 words for each topic. It is noteworthy that the PorterStemmer yields the roots of words, leading to terms like ``uncertainti'', ``valu'', and ``immedi''. Topic 1 primarily centers on risk and uncertainty, topics 2 and 4 delve into opportunity cost and alternative investment options, while topic 3 concerns urgent or immediate needs. We have also created an interactive web-based visualization that allows one to explore the topics discovered by the LDA model, their prominence, and relationships in a more intuitive way. It can be reached at this anonymous link \url{https://mksc-llm.tiiny.site/}. One can for instance click on the word ``risk", and that highlights the topics (topic 1) with the highest association with the word.

\begin{table}[H]
    \centering
    \begin{tabular}{cc||cc||cc||cc}
    \toprule
    \multicolumn{2}{c||}{\textbf{Topic 1}} & \multicolumn{2}{c||}{\textbf{Topic 2}} & \multicolumn{2}{c||}{\textbf{Topic 3}} & \multicolumn{2}{c}{\textbf{Topic 4}} \\
    \midrule
    \midrule
    valu & risk & option & receiv & need & option & money & valu \\
    receiv & option & wait & month & valu & choos & receiv & invest \\
    may & time & choos & time & month & immedi & time & futur \\
    month & futur & use & one & wait & receiv & month & potenti \\
    consid & choos & addit & invest & consid & factor & option & amount \\
    chang & factor & consid & therefor & howev & decis & use & worth \\
    one & prefer & earlier & offer & futur & make & earn & one \\
    wait & chip & may & although & urgent & increas & choos & due \\
    uncertainti & decreas & get & two & depend & financi & opportun & prefer \\
    also & immedi & reason & prefer & time & n't & consid & decis \\
    \bottomrule
    \end{tabular}
    \caption{Top 20 words for each topic.}
    \label{tab:top_words}
\end{table}

\begin{figure}[ht]
    \centering
    \begin{subfigure}{.32\textwidth}
        \centering
        \includegraphics[width=\linewidth]{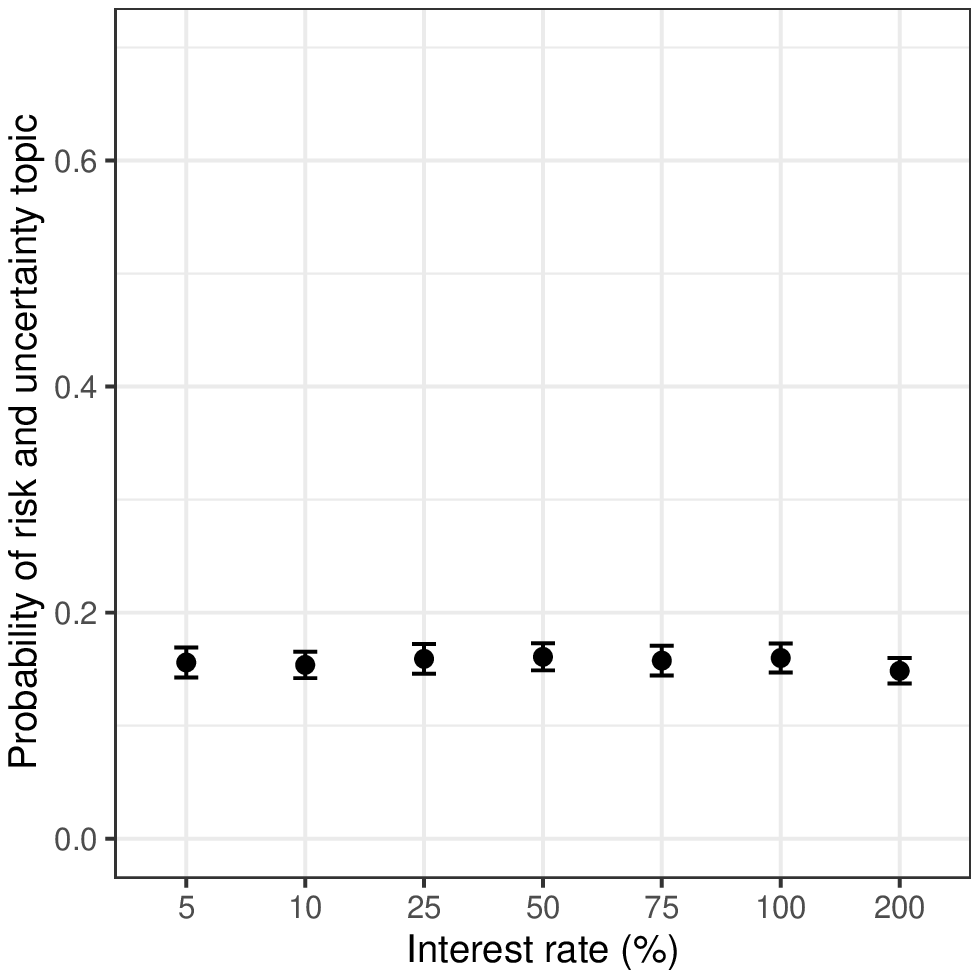}
        \caption{Risk and Uncertainty}
        \label{fig:interest_topics_sub1}
    \end{subfigure}%
    \hfill
    \begin{subfigure}{.32\textwidth}
        \centering
        \includegraphics[width=\linewidth]{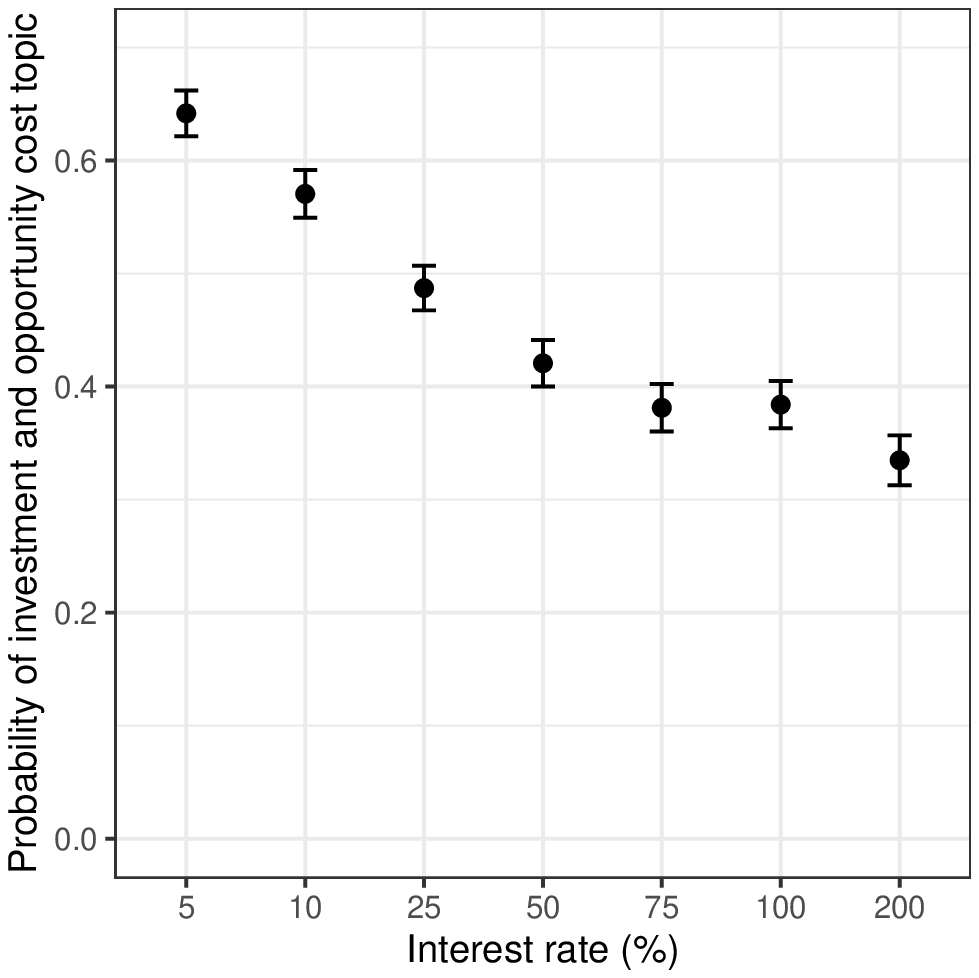}
        \caption{Opportunity cost}
        \label{fig:interest_topics_sub2}
    \end{subfigure}%
    \hfill
    \begin{subfigure}{.32\textwidth}
        \centering
        \includegraphics[width=\linewidth]{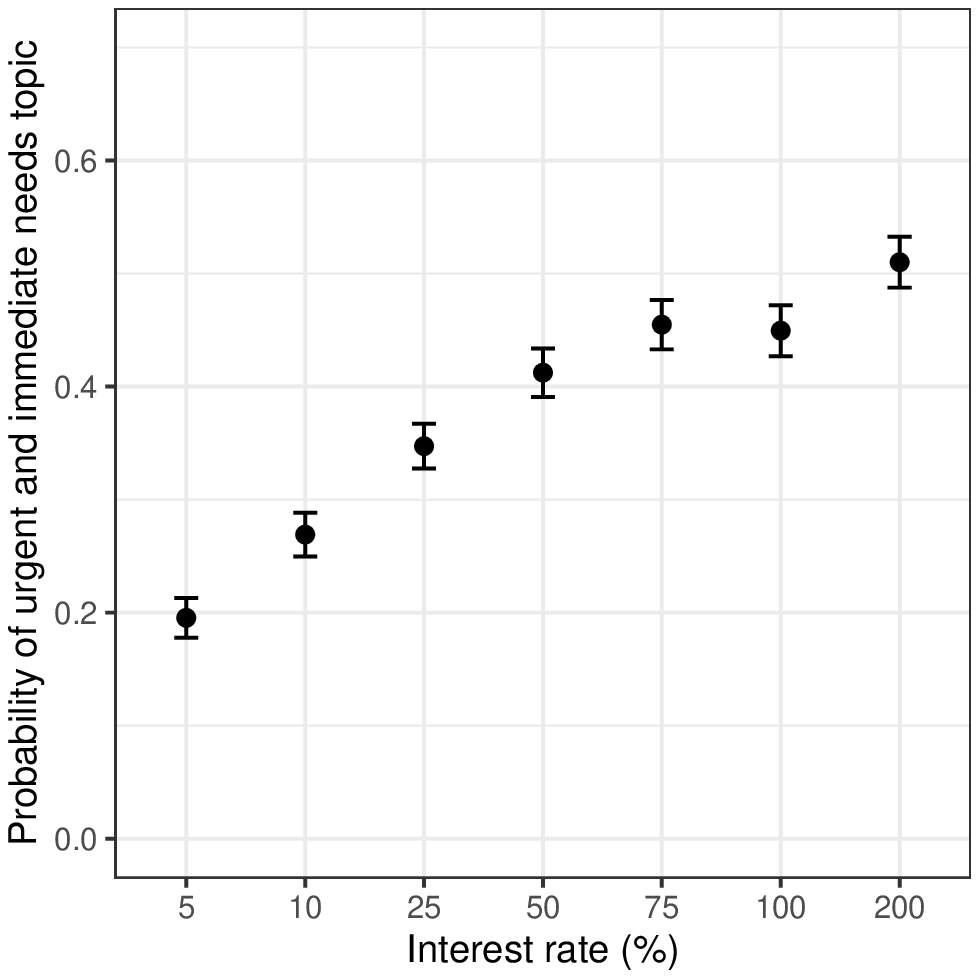}
        \caption{Urgency}
        \label{fig:interest_topics_sub3}
    \end{subfigure}
    \caption{Probability of each topic across different interest conditions. The displayed intervals correspond to the 95\% confidence intervals, clustered at the level of experimental cells (language-delay-interest).}
    \label{fig:interest_topics}
\end{figure}

\begin{figure}[ht]
    \centering
    \begin{subfigure}{.32\textwidth}
        \centering
        \includegraphics[width=\linewidth]{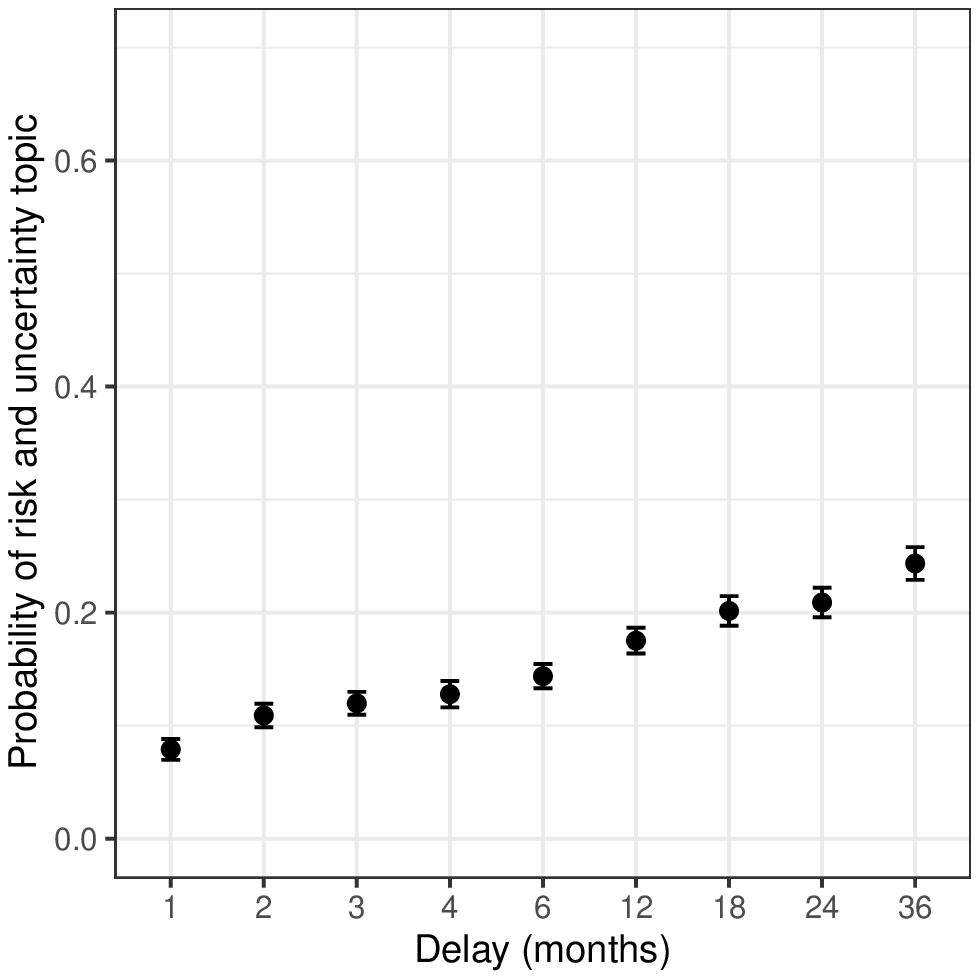}
        \caption{Risk and Uncertainty}
        \label{fig:delay_topics_sub1}
    \end{subfigure}%
    \hfill
    \begin{subfigure}{.32\textwidth}
        \centering
        \includegraphics[width=\linewidth]{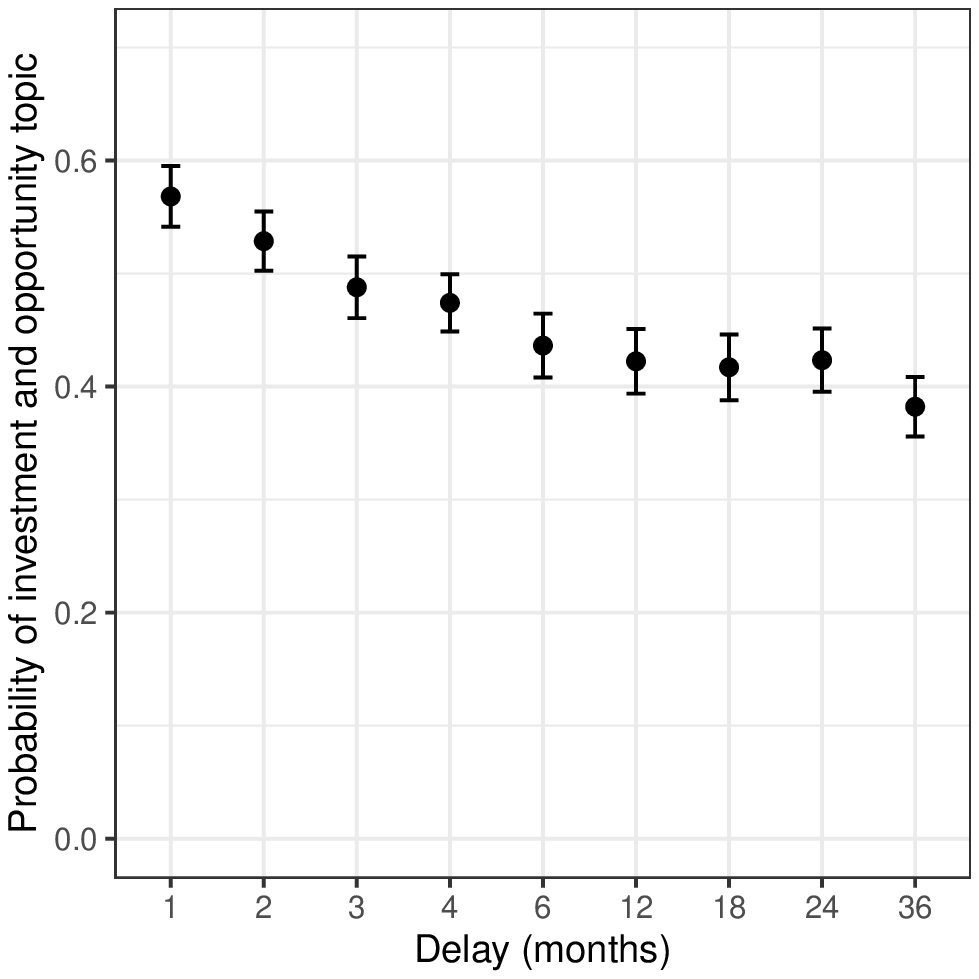}
        \caption{Opportunity cost }
        \label{fig:delay_topics_sub2}
    \end{subfigure}%
    \hfill
    \begin{subfigure}{.32\textwidth}
        \centering
        \includegraphics[width=\linewidth]{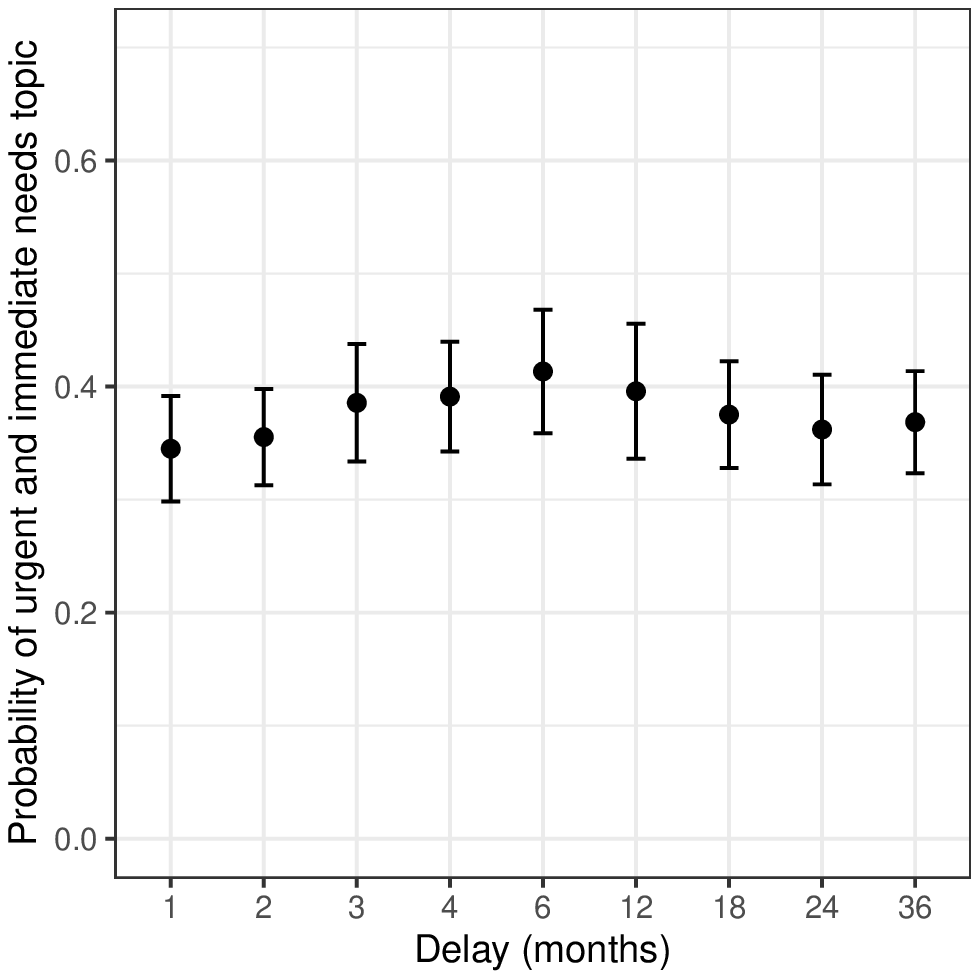}
        \caption{Urgency}
        \label{fig:delay_topics_sub3}
    \end{subfigure}
    \caption{Probability of each topic across different delay conditions. The displayed intervals correspond to the 95\% confidence intervals, clustered at the level of experimental cells (language-delay-interest).}
    \label{fig:delay_topics}
\end{figure}

In the following, we use the label ``risk and uncertainty'' for topic 1. We create an umbrella topic by merging topics 2 and 4 and labeling it ``investment and opportunity cost,'' and we name topic 3 ``urgent and immediate needs.'' Using the trained LDA model, we can assign each of the descriptions generated by GPT-4 to a mix of these three topics. That is, each description will be assigned a vector of three probabilities, reflecting the mix of topics discussed. In Figures~\ref{fig:interest_topics}-\ref{fig:delay_topics}, we display the probability of each topic across different interest and delay conditions. A few interesting patterns emerge. First, when a larger interest is offered (Figure~\ref{fig:interest_topics}), the probability of discussing opportunity cost or alternative investments decreases, with discussions about urgent and immediate needs taking their place. Notably, throughout different ranges of interest rates, the probability of the ``risk and uncertainty'' topic remains flat. Second, as the time delay between the two options increases (Figure~\ref{fig:delay_topics}), the share of the ``risk and uncertainty'' topic goes up, suppressing discussions on opportunity cost without affecting discussions of urgency.

These patterns are intriguing and parallel how human decision-makers might allocate priorities or mental resources across various factors, including urgency, uncertainty, and opportunity cost. For instance, these results can be used to understand the role of risk and uncertainty and why humans have different discount rates over different time horizons~\citep{frederick2002time,mischel2003sustaining}. They can also explain why consumers opt for choices with a higher variety for later periods~\citep{salisbury2008future}. This suggests that even if chain-of-thought conjoint analyses might not be directly useful for eliciting preferences, they could help formulate hypotheses for testing with human subjects, thus aiding in understanding the underpinnings of preferences. The patterns in Figures~\ref{fig:interest_topics}-\ref{fig:delay_topics} can be generated for other decisions or conjoint exercises to propose hypotheses for factors/mediators that might influence decisions.

\begin{figure}[ht]
    \centering
    \begin{subfigure}{.32\textwidth}
        \centering
        \includegraphics[width=\linewidth]{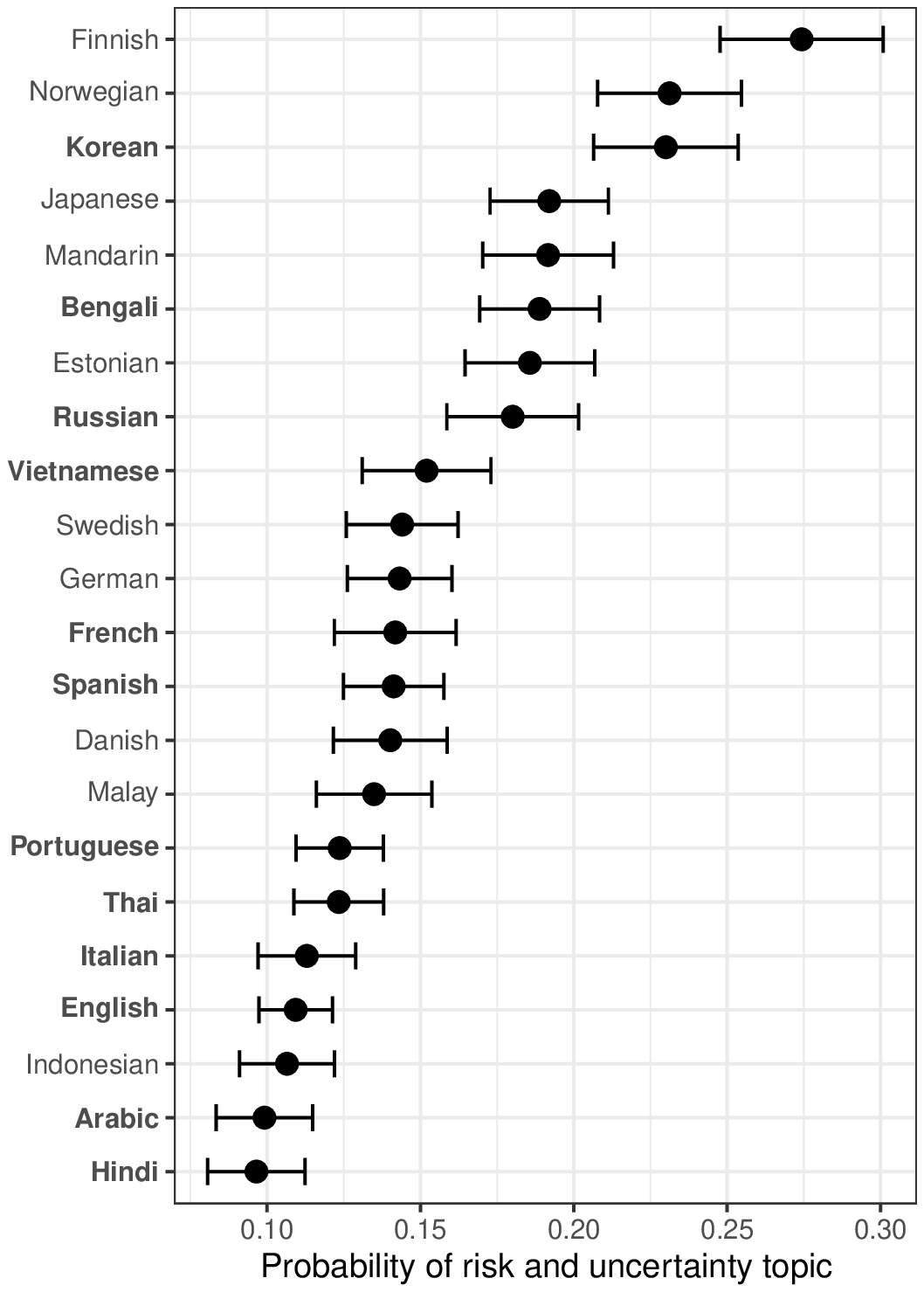}
        \caption{Risk and Uncertainty}
        \label{fig:lang_topics_sub1}
    \end{subfigure}%
    \hfill
    \begin{subfigure}{.32\textwidth}
        \centering
        \includegraphics[width=\linewidth]{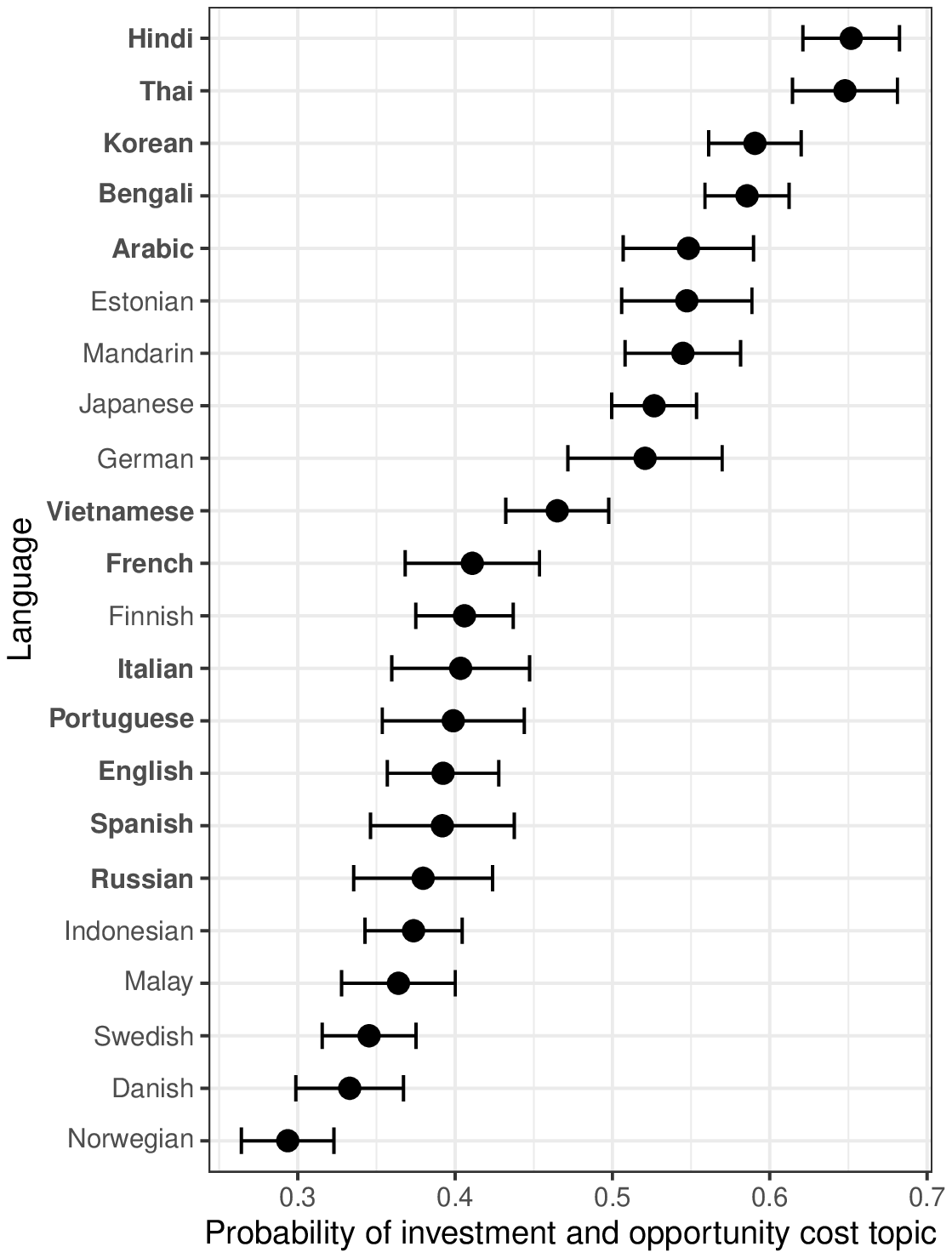}
        \caption{Opportunity cost }
        \label{fig:lang_topics_sub2}
    \end{subfigure}%
    \hfill
    \begin{subfigure}{.32\textwidth}
        \centering
        \includegraphics[width=\linewidth]{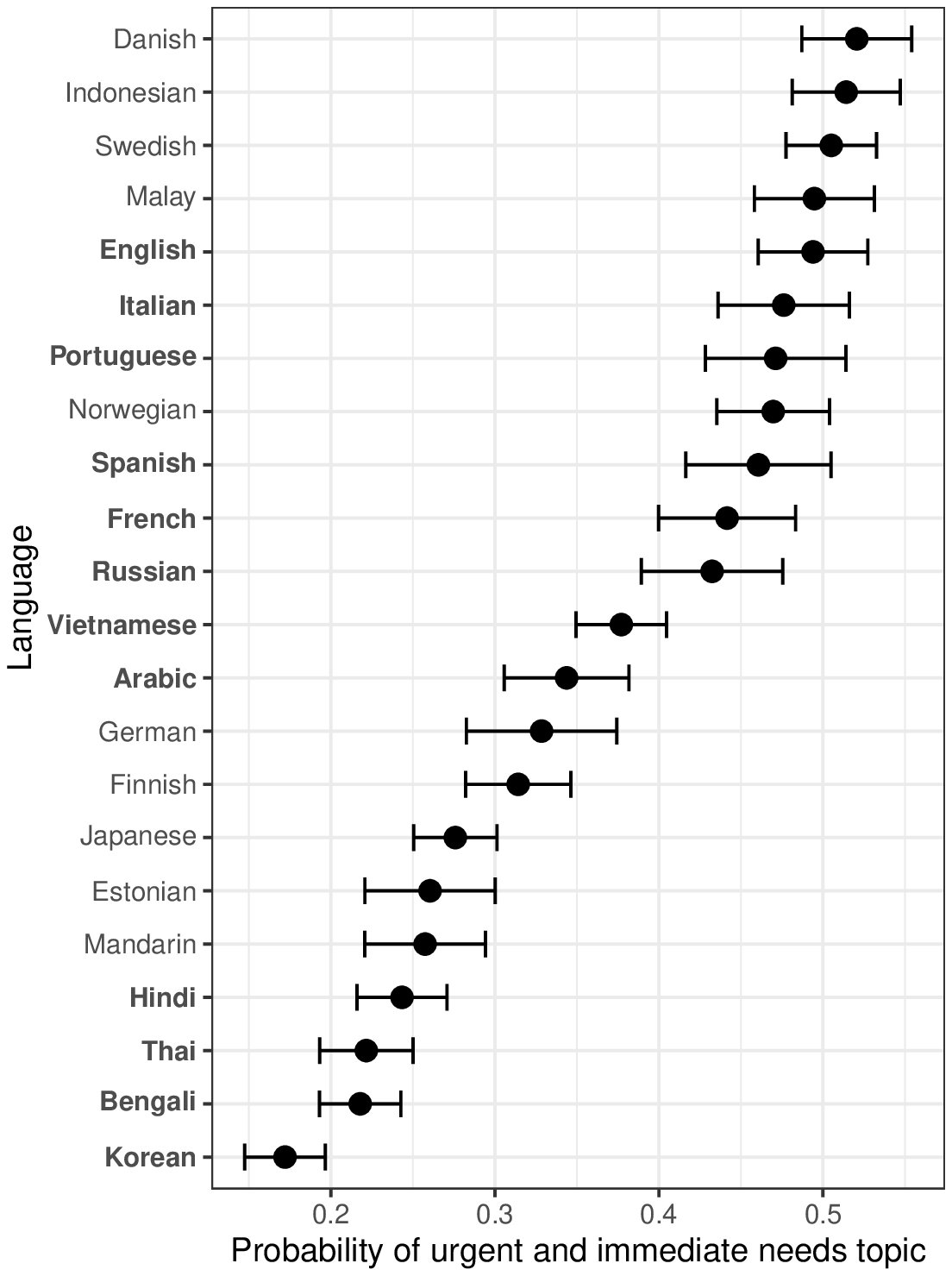}
        \caption{Urgency}
        \label{fig:lang_topics_sub3}
    \end{subfigure}
    \caption{Probability of each topic across different delay conditions. The displayed intervals correspond to the 95\% confidence intervals, clustered at the level of experimental cells (language-delay-interest).  Languages with strong FTR are displayed in bold font. }
    \label{fig:lang_topics}
\end{figure}

We can repeat this exercise across various languages. In Figure~\ref{fig:lang_topics}, we display the propensity of each topic across the 22 languages studied here. Significant heterogeneity is observed across languages. These patterns highlight how GPT discusses or considers different sets of topics across languages in the context of intertemporal choices. Likely, these differences arise from variations in topic co-occurrences across different cultures and languages when discussing intertemporal choice. To demonstrate why this information might be valuable to researchers, let's identify which topic varies the most between weak and strong FTR languages. To achieve this, consider the following specification:
\begin{equation}
\bY_{k} = \alpha \cdot \mathbbm{1}_{\ell_{k} \in \mathcal{S}} + \eta_{d_{k}i_{k}} + \epsilon_k,
\label{eq:topic_main_strong}
\end{equation}
where \( k \) indexes the sample, and \( \ell_k \) is the query language used for drawing sample \( k \). The interest rate and the delay for the experiment condition corresponding to sample \( k \) are represented by \( i_k \) and \( d_k \), respectively. \( \mathcal{S} \) is the set of languages with strong FTR in our dataset. \( \bY_{k} \) is the normalized propensity of a given topic which is calculated as the probability of a given topic in the query response divided by the average probability of that topic across all queries. This helps us normalize for differences in the overall prevalence of different topics, and the coefficient \( \alpha \) can be interpreted as the percentage difference in the prevalence of a given topic in weak relative to strong FTR languages. Finally, \( \eta_{d_k i_k} \) represents the interest-delay fixed effects. By including these fixed effects in the regression, we are essentially comparing different languages (those with strong and weak FTR) for the same set of interest-delay conditions and investigating whether the likelihood of choosing the larger, later option systematically differs between languages with strong and weak FTR.

The results from specification~\eqref{eq:topic_main_strong} for each of the three topics-risk and uncertainty, opportunity cost, and urgency-are displayed in Table~\ref{tab:topic_strong_weak_FTR}. The results indicate that the topic of risk \& uncertainty is discussed 20\% less frequently than average in strong FTR languages. This figure is 13.8\% for opportunity cost and -8.3\% for urgency. This suggests that the difference in intertemporal preferences between strong and weak FTR languages might primarily be due to differences in risk and uncertainty aversion. Interestingly, the difference in risk-taking between weak and strong FTR languages in corporates and banks has been documented in the corporate finance literature~\citep{chen2017languages,osei2021language}.

Our analysis above showcases that such variations might help researchers identify factors or mediators that influence choices across different customer segments or choice contexts. Given the role of language in our perception, we believe our results underscore GPT's potential to pinpoint areas and factors that could explain differences in consumer preferences. However, we do not believe that GPT can be used to directly elicit preferences.

\begin{table}[!htbp] 
\centering 
\caption{Differences in prevalence of topics across strong and weak FTR languages.} 
\label{tab:topic_strong_weak_FTR} 
\resizebox{0.8\textwidth}{!}{ 
\begin{tabular}{@{\extracolsep{5pt}}lccc} 
\\[-1.8ex]\hline 
\hline \\[-1.8ex] 
 & \multicolumn{3}{c}{\textit{Dependent variable: Prevalence of topic (normalized)}} \\ 
\cline{2-4} 
\\[-1.8ex] & Risk \& Uncertainty & Opportunity cost& Urgency \\ 
\\[-1.8ex] & (1) & (2) & (3)\\ 
\hline \\[-1.8ex] 
Strong FTR & $-$0.209$^{***}$ & 0.138$^{***}$ & $-$0.083$^{***}$ \\ 
 & (0.025) & (0.015) & (0.021) \\ 
 & & & \\ 
\hline \\[-1.8ex] 
Observations & 13,860 & 13,860 & 13,860 \\ 
R$^{2}$ & 0.086 & 0.169 & 0.128 \\ 
Adjusted R$^{2}$ & 0.082 & 0.165 & 0.124 \\ 
Residual Std. Error (df = 13796) & 1.146 & 0.600 & 0.762 \\ 
\hline 
\hline \\[-1.8ex] 
\textit{Note:}  & \multicolumn{3}{r}{$^{*}$p$<$0.1; $^{**}$p$<$0.05; $^{***}$p$<$0.01} \\ 
\multicolumn{4}{l}{All regressions include delay-interest fixed effects. } \\
\multicolumn{4}{l}{All standard errors are clustered at  the experimental cell~(language-delay-interest) level. } \\
\end{tabular}}
\end{table}

\section{Discussion and Conclusions}
\label{conclusion}
Arguably, one of the most notable advancements in AI research in recent years has been the emergence of LLMs. Built on the Transformer architecture~\citep{vaswani2017attention}, these models focus on predicting subsequent words, enabling LLMs to generate ``plausible'' text based on extensive training data. LLMs have demonstrated exceptional performance in tasks such as coding, music, and abstraction~\citep{bubeck2023sparks}. The vast size of LLMs could lead to the development of ``neural circuits'' that specialize in certain tasks or reasoning within specific contexts~\citep{liu2022transformers,zhang2022unveiling}. 

Owing to the capacity of LLMs to mimic humans across various contexts and their inherent stochastic nature, social scientists have explored their use in producing survey samples. They achieve this by employing different contexts, such as prompting GPT to emulate responses generated by distinct demographic groups~\citep{Chen2023Emergence,brand2023using,horton2023large,argyle2022out}. These prompts enable LLMs to draw response samples conditional on the given context, like sociodemographic groups. To test this capability, we use GPT-3.5 and 4 to measure intertemporal preferences. The focus on intertemporal preferences is due to (a) the existence of a large body of literature that can be used for benchmarking and understanding properties of such preferences, (b) the literature that documents systematic differences in intertemporal preferences across languages, and (c) the potential of LLMs to recommend products or educate users on products involving intertemporal trade-offs such as mortgages, investments, and insurance services. This potential prompts us to investigate if LLMs are indeed able to make such trade-offs.

In this paper, we document that GPT's choices in intertemporal trade-offs change systematically when it is queried across different models (GPT-3.5 versus 4), different ways to query (chain-of-thought versus standard conjoint), and across languages. We show that GPT-3.5 displays lexicographical preferences when confronted with intertemporal trade-offs, favoring immediate rewards but struggling to effectively trade off smaller, sooner rewards with larger, later ones. This pattern changes for GPT-4, but we still document larger discount rates than those measured among human subjects. We propose an alternative to the standard conjoint with LLMs, which we call ``chain-of-thought conjoint.'' We show that GPT exhibits more patience when queried using the chain-of-thought conjoint approach. Given the extent of changes across different ways of eliciting preferences, we argue that eliciting preferences directly using GPT responses could be misleading. Interestingly, however, we show that by using a combination of topic modeling and chain-of-thought conjoint one can generate a set of topics or mediators that play a role in decision-making and then test them in human subjects. Furthermore, across our experiments, GPT exhibits greater patience in languages with weak FTR, such as German and Mandarin, compared to those with strong FTR, like English and French. These results are consistent with prior literature on the impact of language on intertemporal preferences and suggest that GPT can be a useful tool for understanding how language or other factors might influence consumer behavior. These results indicate that, while using GPT to directly elicit preferences could be misleading, GPT can play an important role in generating hypotheses that can help researchers understand the underpinnings of heterogeneity in preferences across context and customer segments.


\newpage

\pagebreak
\newpage
\section*{Funding and Competing Interests Declaration}
All authors certify that they have no affiliations with or involvement in any organization or entity with any financial interest or non-financial interest in the subject matter or materials discussed in this manuscript. The authors have no funding to report.
\newrefcontext[sorting=nty]
\printbibliography
\newpage

\section*{Online Appendix: Further Investigation of GPT-3.5 Choices}
The findings presented in section~\ref{sec:het_langs} demonstrate that GPT-3.5's choices do not vary as a function of different interest rates offered. This pattern is strange and is not consistent with a discounted utility decision maker or humans ~\citep{andersen2008eliciting}. In this online appendix, we delve deeper into decisions made by GPT-3.5, first by investigating differences across strong and weak FTR languages, and then by formally showing that GPT-3.5 exhibits lexicographic preferences in terms of time and rewards.

\subsection*{Strong versus weak FTR languages (GPT-3.5)}

The observations in Figure~\ref{fig:avg_reward_choice_langs} emphasize the potential role of language in time preferences, consistent with previous findings by \citet{chen2013effect} and others \citep{boroditsky2001does, casasanto2008s,ayres2023languages}. Our results in Figure~\ref{fig:het_interest} show that GPT-3.5 decisions are not consistent with any discounted utility decision models (see above) or decision-making by humans. Therefore, the data from GPT-3.5 decisions are not suitable for estimating a discount rate in a discounted utility model. However, we now examine whether GPT-3.5 is more likely to choose the larger, later reward in languages with strong FTR compared to those with weak FTR. We consider the following specification:

\begin{equation}
\bY_{k} = \alpha \cdot \mathbbm{1}_{\ell_{k} \in \mathcal{S}} + \eta_{d_{k}i_{k}} + \epsilon_k,
\label{eq:main_strong}
\end{equation}
where $k$ indexes the sample, and $\ell_k$ is the query language used for drawing sample $k$. The interest rate and the delay for the experiment condition corresponding to sample $k$ are represented by $i_k$ and $d_k$, respectively. $\mathcal{S}$ is the set of languages with strong FTR in our dataset. $\bY_{k}$ is a dummy variable that equals one if the larger, later option was selected. Finally, $\eta_{d_k i_k}$ represents interest-delay fixed effects. By including these fixed effects in the regression, we are essentially comparing different languages (those with strong and weak FTR) for the same set of interest-delay conditions and investigating whether the likelihood of choosing the larger, later option systematically differs between languages with strong and weak FTR.

We display the estimates for equation~\eqref{eq:main_strong} in Table~\ref{tab:main_strong}. The first column presents the estimates without any fixed effects, and we progressively include additional fixed effects as we proceed to the third column. The standard errors are all clustered at the experimental cell (language-delay-interest) level. Our findings remain consistent as we include more fixed effects. Our preferred specification with Delay-Interest fixed effects shows that GPT-3.5 is 2.8\% less likely to choose the larger, later reward when queried in languages with strong FTR, compared to its responses in languages with weak FTR. These results are consistent with previous studies such as~\citet{chen2013effect,boroditsky2001does,herz2021time}. However, the patterns in Figure~\ref{fig:het_interest} indicate that GPT-3.5's decisions do not vary with changes in the interest rate and remain relatively stable across a wide range of interest rate conditions, varying from 5 to 200\% per year. These patterns are peculiar, as human decision-makers do alter their intertemporal choices when interest rates increase, as evidenced by, for example, see \citet{andersen2008eliciting}.  In the next section, we examine whether GPT-3.5's choices satisfy simple regularity conditions that one would expect human choices to adhere to.

\begin{table}[!htbp] \centering 
\caption{The difference in the propensity of selecting the larger, later option between languages with strong and weak FTR.}
\label{tab:main_strong} 
\resizebox{0.8\textwidth}{!}{ 
\begin{tabular}{@{\extracolsep{5pt}}lccc} 
\\[-1.8ex]\hline 
\hline \\[-1.8ex] 
 & \multicolumn{3}{c}{\textit{Dependent variable:}} \\ 
\cline{2-4} 
\\[-1.8ex] & \multicolumn{3}{c}{Choosing the delayed reward} \\ 
\\[-1.8ex] & (1) & (2) & (3)\\ 
\hline \\[-1.8ex] 
 Strong FTR & $-$0.028$^{***}$ & $-$0.028$^{***}$ & $-$0.028$^{***}$ \\ 
  & (0.009) & (0.007) & (0.007) \\ 
  & & & \\ 
 Constant & 0.232$^{***}$ &  &  \\ 
  & (0.006) &  &  \\ 
  & & & \\ 
\hline \\[-1.8ex] 
Delay FE &  & X & \\ 
Delay-Interest FE &  &  & X \\ 
Observations & 138,600 & 138,600 & 138,600 \\ 
R$^{2}$ & 0.001 & 0.050 & 0.059 \\ 
Adjusted R$^{2}$ & 0.001 & 0.050 & 0.059 \\ 
Residual Std. Error & 0.412 (df = 138598) & 0.402 (df = 138590) & 0.400 (df = 138536) \\ 
\hline 
\hline \\[-1.8ex] 
\textit{Note:}  & \multicolumn{3}{r}{$^{*}$p$<$0.1; $^{**}$p$<$0.05; $^{***}$p$<$0.01} \\ 
\multicolumn{4}{l}{All standard errors are clustered at  the experimental cell~(language-delay-interest) level. } \\
\end{tabular}}
\end{table}

\subsection*{Are GPT-3.5 choices consistent with a ``proper'' choice model?}
\label{sec:proper}

The evidence documented in sections~\ref{sec:het_langs} and above highlights notable heterogeneity in GPT-3.5 choices across languages. Our findings in section~\ref{sec:het_interest}, however, suggest that the decision patterns do not seem to conform to an exponentially discounted utility model since the likelihood of selecting the larger, later reward changes in the interest rate. These patterns deviate from human decision-making behavior. Motivated by this observation, in this section, we first establish a set of straightforward regularity conditions that various models of intertemporal choices, such as exponential, hyperbolic, or quasi-hyperbolic models, fulfill. Next, we assess whether GPT-3.5 choices adhere to these regularity conditions.

We start by defining a ``proper'' random utility function over tuples $(t, r)$, where $t$ represents the time a reward is received, and $r$ denotes the value of the reward.
\begin{definition} 
A random utility function $U:(\mathcal{T},\mathcal{R}) \rightarrow  \mathds{R}$ is ``proper'' if for $\forall t_1, t_2, r_1$ the probability $v(r_2) = P\left(U(t_2, r_2) > U(t_1, r_1)\right)$ is an increasing function of $r_2$.
\end{definition}

A proper random utility function essentially represents a utility function where the probability of choosing the larger, later reward increases as the size of the later reward grows. This condition is satisfied by exponential, hyperbolic, quasi-hyperbolic, or any other standard discounted utility model. This essentially implies that choosing the larger, later reward is positively correlated with the gap in rewards, or in other words, $cor(\mathbbm{1}_{\{U(t_2, r_2) > U(t_1, r_1)\}} , r_2 - r_1) > 0$. We now examine whether GPT choices indeed fulfill this condition using the following specification:

\begin{equation}
\bY_{k} = \alpha \cdot (r_2 - r_1) + \eta_{\ell_{k} d_k} + \epsilon_k,
\label{eq:proper_test}
\end{equation}

where $k$ indexes the sample, and $\ell_k$ is the query language used for drawing sample $k$. The delay for the experiment condition corresponding to sample $k$ is denoted by $d_k$. $\bY_{k}$ is a dummy variable that equals one if the larger, later option was selected. $r_2 - r_1$ is the difference between the rewards, measured in increments of 1000 tokens. Finally, $\eta_{\ell_{k} d_k}$ represents language-delay fixed effects. By including these fixed effects in the regression, we are comparing different interest rate  conditions while holding language and delay fixed.

Table~\ref{tab:proper_test} presents the estimates from equation~\eqref{eq:proper_test}. The first column displays the estimate for the regression without any fixed effects. The coefficient is slightly negative and statistically significant. Note that we expect this coefficient to be positive, as increasing $r_2$ should make choosing the delayed reward option more likely. In column~(2), we incorporate language fixed effects, but the coefficient remains unchanged. Finally, in column~(3), we add language-delay fixed effects, and  the coefficient shrinks in size and becomes statistically insignificant. This analysis suggests that GPT-3.5 choices cannot be rationalized by any proper utility function. It is important to emphasize once again that these results demonstrate GPT-3.5's inability to reasonably trade off larger, later rewards with smaller, sooner ones.

\begin{table}[!htbp] \centering 
\caption{Testing if GPT-3.5 choices are ``proper''. Our results show that increasing the larger, later reward does not increase the likelihood of GPT-3.5 choosing the larger, later reward. Data for this study were sourced from GPT-3.5, employing the experimental design described in section~\ref{sec:exp_design}.}
\label{tab:proper_test} 
\resizebox{0.8\textwidth}{!}{ 
\begin{tabular}{@{\extracolsep{5pt}}lccc} 
\\[-1.8ex]\hline 
\hline \\[-1.8ex] 
 & \multicolumn{3}{c}{\textit{Dependent variable:}} \\ 
\cline{2-4} 
\\[-1.8ex] & \multicolumn{3}{c}{Choosing the delayed reward} \\ 
\\[-1.8ex] & (1) & (2) & (3)\\ 
\hline \\[-1.8ex] 
 Difference in rewards (in 1000 tokens) & $-$0.009$^{***}$ & $-$0.009$^{***}$ & 0.001 \\ 
  & (0.001) & (0.001) & (0.001) \\ 
  & & & \\ 
 Constant & 0.227$^{***}$ &  &  \\ 
  & (0.004) &  &  \\ 
  & & & \\ 
\hline \\[-1.8ex] 
Language FE &  & X &  \\ 
Language-Delay FE &  &  & X \\ 
Observations & 138,600 & 138,600 & 138,600 \\ 
R$^{2}$ & 0.006 & 0.048 & 0.114 \\ 
Adjusted R$^{2}$ & 0.006 & 0.048 & 0.112 \\ 
Residual Std. Error & 0.411 (df = 138598) & 0.402 (df = 138577) & 0.388 (df = 138401) \\ 
\hline 
\hline \\[-1.8ex] 
\textit{Note:}  & \multicolumn{3}{r}{$^{*}$p$<$0.1; $^{**}$p$<$0.05; $^{***}$p$<$0.01} \\ 
\multicolumn{4}{l}{All standard errors are clustered at  the experimental cell~(language-delay-interest) level. } \\
\end{tabular}}
\end{table}

\subsubsection*{Are GPT-3.5 choices consistent with lexicographic preferences over time and rewards?}

The findings above suggest that GPT-3.5 demonstrates lexicographic preferences in relation to time and rewards, essentially ignoring reward differences when the time periods are not the same. To examine this, we prompt GPT-3.5 across the 22 languages in our dataset, but this time we ask GPT-3.5 to choose between tuples $(t, r_1)$ and $(t, r_2)$, where $t$ is part of the set ${2, 3, 4, 5, 7, 13, 25}$, $r_1 = 1000$ tokens, and $r_2$ belongs to the set ${1041, 1082, 1401, 1781, 3174, 5061, 7376}$. In total, we have $22 \times 7 \times 7 = 1078$ experimental cells and gather 100 samples for each condition. Note that these choices are essentially trivial, asking GPT-3.5, for example, whether it prefers 1041 tokens in 2 months or 1000 tokens in 2 months from now.

\begin{figure}[H]
    \centering
    \includegraphics[width = 0.8\textwidth]{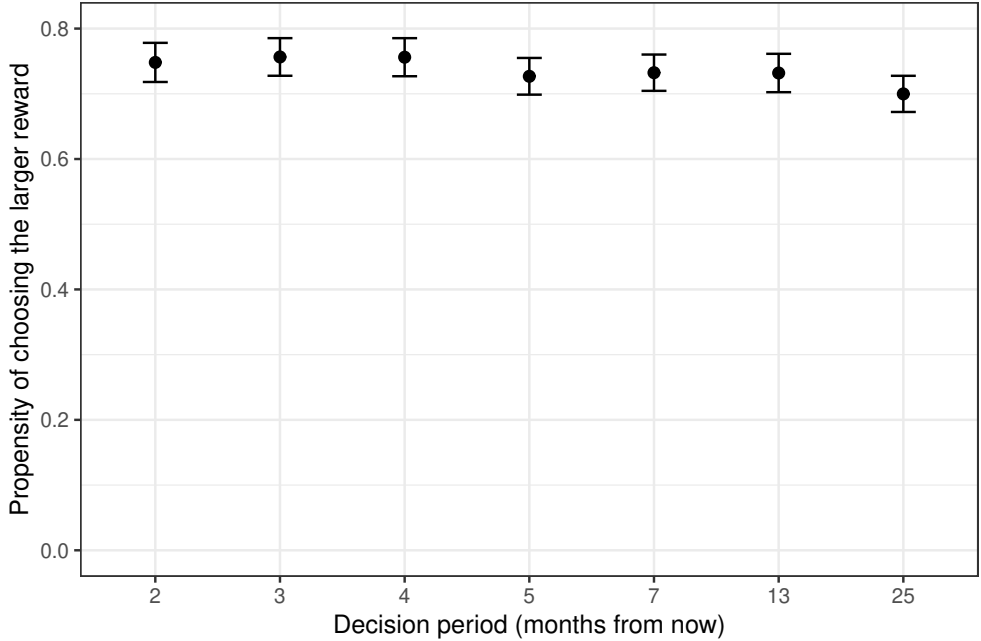}
\caption{Probability of selecting the larger option when both choices are offered $t$ months from the present.}
\label{fig:same_period_choice}
\end{figure}

We plot the likelihood of choosing the larger option across different in Figure~\ref{fig:same_period_choice}. The average probability of selecting the larger option is equal to $74\%$, which means that GPT-3.5 is able to compare the options when they are presented in the same time frame. 

Comparing the average 22\% probability of selecting the larger later reward when there is a delay between time periods (as shown in Figure~\ref{fig:avg_reward_choice_langs}) with these results suggests that GPT-3.5 is indeed exhibiting lexicographic preferences. However, it is important to note that the choices mentioned above appear trivial; GPT-3.5 should pick the higher reward with a probability of 1, given that both rewards are offered in the same time period $t$. Yet, this is not what we observe due to inherent randomness in the choice. Under a standard lexicographic utility function, the preferences should be deterministic~\citep{kohli2007representation}. To address this, we now formally define a lexicographic random utility model of rewards and the time at which they are received $(t,r)$, and present additional evidence supporting the fact that GPT-3.5 demonstrates lexicographic preferences (in the random utility sense).

\begin{definition}
A random utility function $U:(\mathcal{T},\mathcal{R}) \rightarrow  \mathds{R}$ is lexicographical if for $\forall t_1 \neq t_2$ and $\forall r_1, r_2$ the random variable $\mathbbm{1}_{\{U(t_1, r_1) > U(t_2, r_2)\}}\perp r_1, r_2$, but for $\forall t$, we have $\mathbbm{1}_{\{U(t, r_1) > U(t, r_2)\}} \not\perp r_1, r_2$.
\label{def:lexicographic}
\end{definition}

Our results in Table~\ref{tab:proper_test} indicate that when $t_1 \neq t_2$, $r_2 - r_1 \perp \mathbbm{1}_{\{U(t_1, r_1) > U(t_2, r_2)\}}$, which aligns with the first part of Definition~\ref{def:lexicographic}. We now illustrate that when the time periods are the same ($t_1 = t_2$), GPT-3.5 does systematically shift its decisions as the gap in rewards increases, i.e., $cor(r_2 - r_1, \mathbbm{1}_{\{U(t, r_1) > U(t, r_2)\}}) > 0$. To investigate this, we re-estimate specification~\eqref{eq:proper_test} to examine if GPT-3.5 choices shift as a function of differences in rewards when $t = t_1 = t_2$. The results of this analysis are presented in Table~\ref{tab:lexico_test} and demonstrate that the probability of choosing the larger option systematically changes as the gap between the values increases. This finding indicates that GPT-3.5 is capable of discerning between options' reward sizes when they are presented within the same time frame.

\begin{table}[!htbp] \centering 
  \caption{The effect of changing the reward size on the probability of choosing the option with the larger reward. This study involves choices that are presented in the same time frame, that is $t_1 = t_2$. The results show that GPT-3.5 is more likely to pick the large options and the gap in reward increases. } 
  \label{tab:lexico_test} 
\resizebox{0.8\textwidth}{!}{ 
\begin{tabular}{@{\extracolsep{5pt}}lccc} 
\\[-1.8ex]\hline 
\hline \\[-1.8ex] 
 & \multicolumn{3}{c}{\textit{Dependent variable:}} \\ 
\cline{2-4} 
\\[-1.8ex] & \multicolumn{3}{c}{Choosing the larger reward} \\ 
\\[-1.8ex] & (1) & (2) & (3)\\ 
\hline \\[-1.8ex] 
Difference in rewards (in 1000 tokens) & 0.012$^{***}$ & 0.012$^{***}$ & 0.012$^{***}$ \\ 
  & (0.002) & (0.002) & (0.002) \\ 
  & & & \\ 
 Constant & 0.712$^{***}$ &  &  \\ 
  & (0.010) &  &  \\ 
  & & & \\ 
\hline \\[-1.8ex] 
Language FE &  & X &  \\ 
Language-Delay FE &  &  & X \\ 
Observations & 107,800 & 107,800 & 107,800 \\ 
R$^{2}$ & 0.004 & 0.092 & 0.110 \\ 
Adjusted R$^{2}$ & 0.004 & 0.092 & 0.108 \\ 
Residual Std. Error & 0.440 (df = 107798) & 0.420 (df = 107777) & 0.416 (df = 107645) \\ 
\hline 
\hline \\[-1.8ex] 
\textit{Note:}  & \multicolumn{3}{r}{$^{*}$p$<$0.1; $^{**}$p$<$0.05; $^{***}$p$<$0.01} \\ 
\multicolumn{4}{l}{All standard errors are clustered at  the experimental cell~(language-delay-interest) level. } \\
\end{tabular}}
\end{table}

\end{document}